\documentclass{article}

\usepackage{microtype}
\usepackage{graphicx}
\usepackage{subfigure}
\usepackage{booktabs} 

\usepackage{amsmath}
\usepackage{amsfonts}
\usepackage{amssymb}

\usepackage{algorithm}
\usepackage{algorithmic}

\usepackage{multirow}
\usepackage{multicol}

\usepackage{stackengine}

\usepackage{hyperref}

\usepackage[accepted]{icml2020}

\icmltitlerunning{Intrinsic Reward Driven Imitation Learning via Generative Model}

\begin{document}

\twocolumn[
\icmltitle{Intrinsic Reward Driven Imitation Learning via Generative Model}




\begin{icmlauthorlist}
\icmlauthor{Xingrui Yu}{uts}
\icmlauthor{Yueming Lyu}{uts}
\icmlauthor{Ivor W. Tsang}{uts}

\end{icmlauthorlist}

\icmlaffiliation{uts}{Australian Artificial Intelligence Institute, University of Technology Sydney}
\icmlcorrespondingauthor{Xingrui Yu}{Xingrui.Yu@student.uts.edu.au}
\icmlcorrespondingauthor{Ivor W. Tsang}{Ivor.Tsang@uts.edu.au}

\icmlkeywords{Machine Learning, Imitation Learning, Inverse Reinforcement Learning, Atari, ICML}

\vskip 0.3in
]



\printAffiliationsAndNotice{}  

\begin{abstract}
Imitation learning in a high-dimensional environment is challenging. Most inverse reinforcement learning (IRL) methods fail to outperform the demonstrator in such a high-dimensional environment, e.g., Atari domain. To address this challenge, we propose a novel reward learning module to generate intrinsic reward signals via a generative model. Our generative method can perform better forward state transition and backward action encoding, which improves the module's dynamics modeling ability in the environment. Thus, our module provides the imitation agent both the intrinsic intention of the demonstrator and a better exploration ability, which is critical for the agent to outperform the demonstrator. Empirical results show that our method outperforms state-of-the-art IRL methods on multiple Atari games, even with one-life demonstration. Remarkably, our method achieves performance that is up to 5 times the performance of the demonstration.

\end{abstract}

\section{Introduction}
\label{sect:introduction}
Imitation Learning (IL) offers an approach to train an agent to mimic the demonstration of an expert. Behavioral cloning (BC) is probably the simplest form of imitation learning \cite{pomerleau1991efficient}. The promise of this method is to train a policy to predict the demonstrator's actions from the states using supervised learning. However, despite its simplicity, behavioral cloning suffers from a compounding error problem if the data distribution diverges too much from the training set \cite{ross2011reduction}. In other words, an initial minor error can result in severe deviation from the demonstrator's behavior. On the other hand, inverse reinforcement learning (IRL) \cite{abbeel2004apprenticeship,ng2000algorithms} aims at recovering a reward function from the demonstration, and then execute reinforcement learning (RL) on that reward function. However, even with many demonstrations, most state-of-the-art inverse reinforcement learning methods fail to outperform the demonstrator in high-dimensional environments, e.g., Atari domain. 

\begin{figure}[t]
    \centering
    \includegraphics[width=0.48\textwidth]{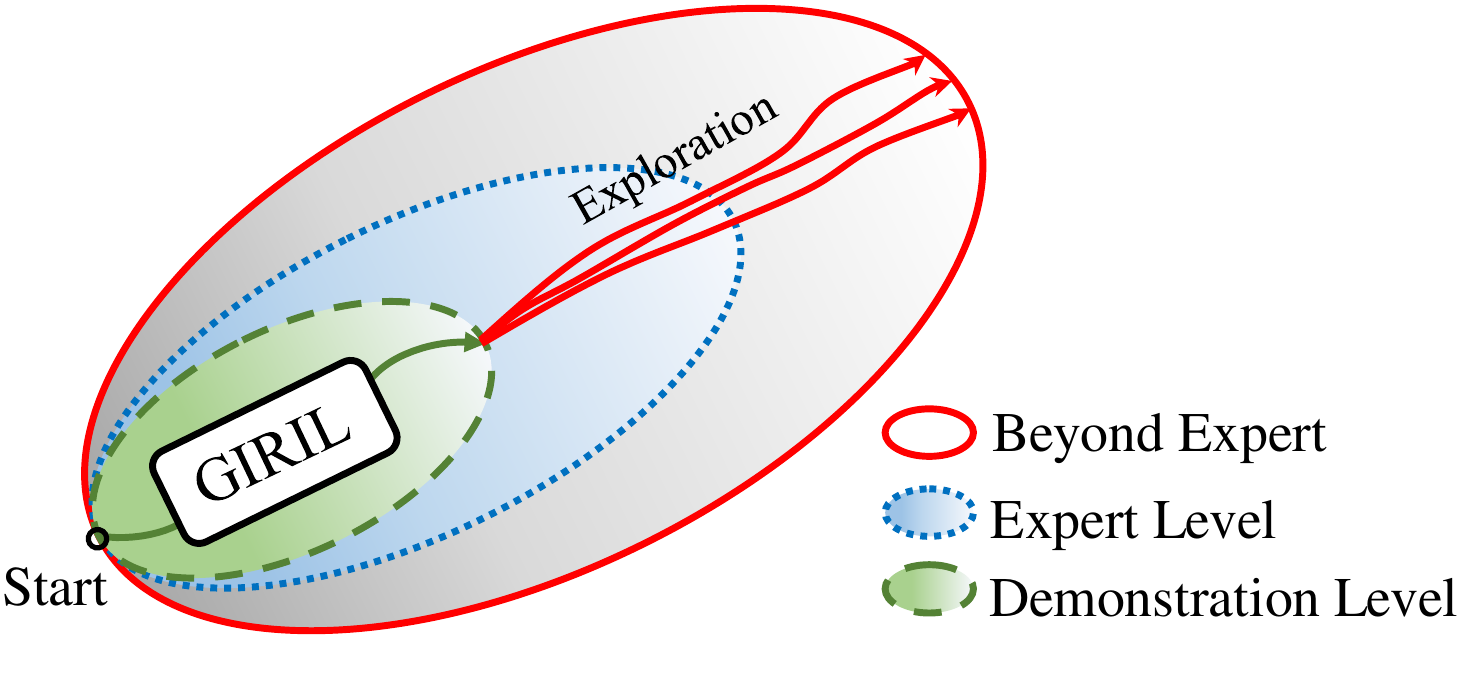}
    \caption{Generative intrinsic reward driven imitation learning (GIRIL) seeks a reward function to achieve three imitation goals. 1) Match the basic demonstration-level performance. 2) Reach the expert-level performance. and 3) Exceed expert-level performance.  GIRIL performs beyond the expert by generating a family of intrinsic rewards for sampling-based self-supervised exploration.} 
    \label{fig:core}
\end{figure}

In our quest to find a solution to this curse of dimensionality, the primary goal to train an agent to reach the expert-level performance with  limited demonstration data. This goal is very challenging  because, in our context, limited demonstration data means that the agent can only learn from the states and actions recorded from gameplay up until the demonstrator, i.e. the player, loses their first life. We call this a ``one-life demonstration". With currently-available approaches, it is unrealistic to expect an agent to even develop sufficient skills to match a player's basic demonstration-level performance. Our ultimate goal for imitation learning is to build an agent that yields better-than-expert imitation performance from only a one-life demonstration. 
Figure \ref{fig:core} illustrates the three steps towards achieving this goal.

Most existing IRL methods fail to reach the first goal in Figure \ref{fig:core}, i.e. the demonstration-level performance. This is because IRL methods seek a reward function that justifies demonstrations only. Given extremely limited state-action data in a one-life demonstration, the recovered reward can be biased. Executing RL on such reward usually results in an agent performing worse than the demonstration. To address this problem, we propose \textit{Generative Intrinsic Reward driven Imitation Learning} (GIRIL), which seeks a family of intrinsic reward functions that enables the agent to do sampling-based self-supervised exploration in the environment. This is critical to better-than-expert performance. 

GIRIL operates by reward inference and policy optimization, and includes a novel generative intrinsic reward learning (GIRL) module based on a generative model. We chose variational autoencoder (VAE) \cite{kingma2013auto} as our model base. It operates by modeling the forward dynamics as a conditional \textit{decoder} and the backward action encoding as a conditional \textit{encoder}. The \textit{decoder} learns to generate diverse future states from the action conditioned on the current state. Accordingly, the \textit{encoder} learns to encode the future state back to the action latent variable (conditioned on the current state). In this way, our generative model performs better forward state transition and backward action encoding, which improves its dynamics modeling ability in the environment. Our model generates a family of intrinsic rewards, which enables the agent to do sampling-based self-supervised exploration in the environment, which is the key to better-than-expert performance.

Within our chosen domain of Atari games, we first generated a one-life demonstration for each of six games and then trained our reward learning module on the corresponding demonstration data. Finally, we optimize the policy on the intrinsic reward that is provided by the learned reward module. Empirical results show that our method, GIRIL, outperforms several state-of-the-art IRL methods on multiple Atari games. Moreover, GIRIL produced agents that exceeded the expert performance for all six games and the one-life demonstration performance by up to 5 times. The implementation will be available online\footnote{https://github.com/xingruiyu/GIRIL}.

\section{Problem Definition and Related Work}
\label{sect:related_work}

\textbf{Problem definition} The problem is formulated as a Markove Decision Process (MDP) defined by a tuple $(\mathcal{S}, \mathcal{A}, P, r, \gamma)$, where $\mathcal{S}$ is the set of states, $\mathcal{A}$ is the set of actions, $P: \mathcal{S}\times\mathcal{A}\times\mathcal{S} \rightarrow \mathbb{R}_{+}$ is the environment transition distribution, $r: \mathcal{S} \rightarrow \mathbb{R}$ is the reward function, and $\gamma \in (0, 1)$ is the discount factor \cite{puterman2014markov}. The expected discounted return of the policy $\pi$ is given by
\begin{equation}\nonumber
    \eta(\pi)=\mathbb{E}_\tau \bigg [\sum_{t=0} \gamma^t r_t \bigg ],
\end{equation}
\noindent where $\tau=(s_0, a_0, \cdots, a_{T-1}, s_T)$ denotes the trajectory, $s_0 \sim \mathbb{P}_0(s_0)$, $a_t \sim \pi(a_t|s_t)$, and $s_{t+1}\sim P(s_{t+1}|s_t, a_t)$.

\textbf{Inverse reinforcement learning} was proposed as a way to find a reward function that could explain observed behavior \cite{ng2000algorithms}. With such a reward function, an optimal policy can be learned via reinforcement learning \cite{sutton1998introduction} techniques. In the maximum entropy variant of inverse reinforcement learning, the aim is to find a reward function that makes the demonstrations appear near-optimal on the principle of maximum entropy \cite{ziebart2008maximum,ziebart2010modeling,boularias2011relative,finn2016guided}. However,  these  learning methods still seek a reward function that justifies the demonstration data only. And, since the demonstration data obviously does not contain information on how to be better than itself, achieving better-than-expert performance with these methods is difficult.

Generative Adversarial Imitation Learning (GAIL) \cite{ho2016generative} treats imitation learning problem as a distribution matching based generative model, which extends IRL by integrating adversarial training technique \cite{NIPS2014_5423}. However, this also means that GAIL inherits some problems from adversarial training along with its benefits, such as instability in the training process. GAIL performs well in low-dimensional application, e.g., MuJoCo. However, it does not scale well to high-dimensional scenarios, such as Atari games \cite{brown2019extrapolating}. Variational adversarial imitation learning (VAIL) \cite{peng2018variational} improves GAIL by compressing the information via variational information bottleneck. We have included both these methods as comparisons to our GIRIL in our experiments. 

\citet{schroecker2019generative} proposed to match the predecessor state-action distributions modeled by the masked autoregressive flows (MAFs) \cite{papamakarios2017masked}. Although they have demonstrated the advantages of their approach against GAIL and BC with robot experiments, their approach still requires multiple demonstrations to reach good performance levels and high-dimensional game environments were not included in their evaluations. 

Despite their generative ability, GAIL and VAIL just match the state-action pairs from the demonstrations only. GIRIL also uses a generative model, but it does not depend on distribution matching. Rather, it simply improves the modeling of both forward dynamics and backward action encoding of MDP in the environment. Moreover, our method utilizes the generative model based reward learning to generate a family of intrinsic rewards for better exploration in the environment, which is critical for better-than-expert imitation.

\textbf{Reward learning based on curiosity}
Beyond the aforementioned methods, reward learning is another important component of reinforcement learning research. For example, intrinsic curiosity module (ICM) is a state-of-the-art exploration algorithm for reward learning \cite{pathak2017curiosity,burda2018large}. ICM transforms high dimensional states into a visual feature space and imposes cross-entropy and Euclidean loss to learn the feature with a self-supervised inverse dynamics model. The prediction error in the feature space becomes the intrinsic reward function for exploration. 
Although ICM has a tendency toward over-fitting, we believe it has potential as a good reward learning module and so have incorporated it to perform that function in the experiments. 
Accordingly, we have treated the resulting algorithm \textit{curiosity-driven imitation learning} (CDIL) as related baseline in our experiments. 

\textbf{Reward learning for video games}
Most imitation learning methods have only been evaluated on low-dimensional environments, and do not scale well to high-dimensional tasks such as video games (e.g., Atari) \cite{ho2016generative,finn2016guided,fu2017learning,qureshi2018adversarial}.  \citet{tucker2018inverse} showed that state-of-the-art IRL methods are unsuccessful on the Atari environments. \citet{hester2018deep} proposed deep Q-learning from demonstrations (DQfD), utilizing demonstrations to accelerate the policy learning in reinforcement learning. 
Since DQfD still requires the ground-true reward for policy learning, it cannot be considered as a pure imitation learning algorithm.
\citet{ibarz2018reward} proposed to learn to play Atari games by combining DQfD and active preference learning \cite{christiano2017deep}. However, it often performs worse than the demonstrator even with thousands of active queries from an oracle. \citet{brown2019extrapolating} learns to extrapolate beyond the sub-optimal demonstrations from observations via IRL. However, their method relies on multiple demonstrations with additional ranking information. Our method outperforms the expert and state-of-the-art IRL methods on multiple Atari games, requiring only a one-life demonstration for each game. Moreover, our method does not require any ground-truth rewards, queries or ranking information. 



\section{Imitation Learning via Generative Model}
\label{sect:method}


Most of existing IRL methods do not scale to high-dimensional space, and IL from a one-life demonstration is even more challenging. Our solution to address this challenge, \textit{Generative Intrinsic Reward driven Imitation Learning} (GIRIL), seeks a reward function that will incentivize the agent to outperform the performance of the expert via sampling-based self-supervised exploration. Our motivation is that the expert player's intentions can be distilled even from extremely limited demonstrations, like the amount of data collected prior to the player losing a single life in an Atari game. And when a notion of the player's intentions is combined with an incentive for further exploration in the environment, the agent should be compelled to meet and exceed each of the three goals: par basic demonstration performance, par expert performance and, ultimately, better-than-expert performance. GIRIL works through two main mechanisms: intrinsic reward inference and policy optimization\footnote{Policy can be optimized with any policy gradient method.}.

\textbf{Intrinsic Rewards}
Since hand-engineered extrinsic rewards are infeasible in complex environments, our solution is to leverage intrinsic rewards that enable the agent to explore actions that reduce the uncertainty in predicting the consequence of the states.
Note, in ICM \cite{pathak2017curiosity}, a network-based regression is used to fit the demonstration data, it is likely to overfit to the limited state-action data drawn from a one-life demonstration.  
And ICM only produces deterministic rewards. 
These two problems limit its exploration ability. Empirically, CDIL can only outperform the basic one-life demonstration  with the guide of limited exploration ability. Therefore, it is imperative to call for intrinsic reward inference with more powerful exploration ability to  achieve better-than-expert performance.

\begin{figure}[t]
    \centering
    \includegraphics[scale=0.27]{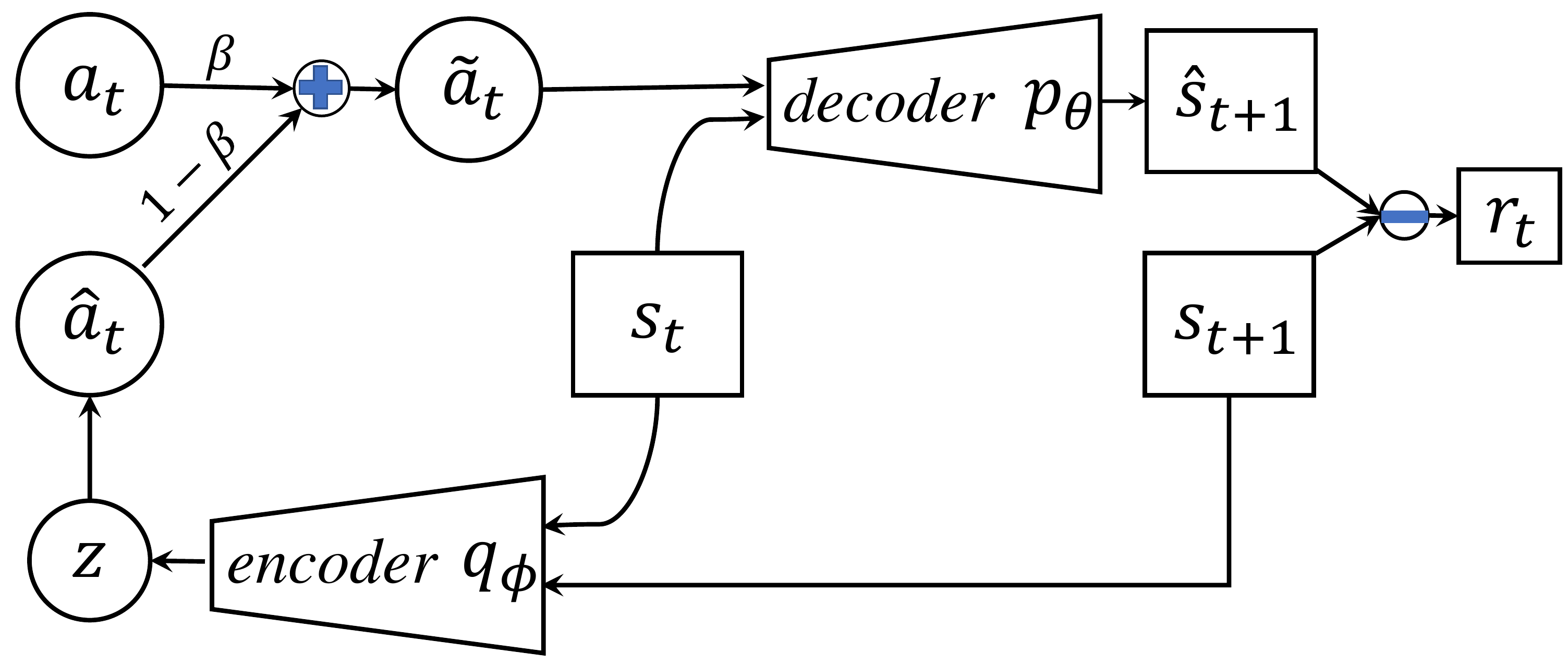}
    \caption{Illustration of the intrinsic reward inference procedure of the proposed GIRIL model.} 
    \label{fig:gir}
\end{figure}

\textbf{\textit{Generative Intrinsic Reward Learning} (GIRL)} To empower the generalization of intrinsic rewards on unseen state-action pairs, our novel reward learning module is based on conditional VAE \cite{sohn2015learning}. As illustrated in Figure \ref{fig:gir}, the module is composed of several neural networks, including recognition network $q_\phi(z|s_t, s_{t+1})$, a generative network $p_\theta(s_{t+1}|z, s_t)$, and prior network $p_\theta(z|s_t)$. We refer to the recognition network (i.e. the probabilistic \textit{encoder}) as a backward action encoding model, and the generative network (i.e. the probabilistic \textit{decoder}) as a forward dynamics model. Maximizing the following objective to optimize the module:
\begin{equation}
\label{eq:gir_obj}
  \begin{aligned}
  \mathcal{L}(s_t, s_{t+1}; \theta, \phi) & = \mathbb{E}_{q_\phi(z|s_t, s_{t+1})}[\log{p_\theta(s_{t+1}|z, s_t)}] \\
    & - {\rm KL}(q_\phi(z|s_t, s_{t+1})\|p_\theta(z|s_t)) \\
    & - \alpha {\rm KL}(q_\phi(\hat{a}_t|s_t, s_{t+1})\|\pi_E(a_t|s_t)) ]
  \end{aligned}
\end{equation}
\noindent where $z$ is the latent variable, $\pi_E(a_t|s_t)$ is the expert policy distribution, $\hat{a}_t=\textrm{Softmax}(z)$ is the transformed latent variable, $\alpha$ is a positive scaling weight.

The first two terms of the right-hand side (RHS) in Eq. (\ref{eq:gir_obj}) denote the evidence lower bound (ELBO) of the conditional VAE \cite{sohn2015learning}. These two terms are critical for our reward learning module to perform better forward state transition and backward action encoding. In other words, the \textit{decoder} performs forward state transition by taking the action $\tilde{a}_t$ and output the reconstruction $\hat{s}_{t+1}$. On the other hand, the \textit{encoder} performs backward action encoding by taking in the states $s_t$ and $s_{t+1}$ and producing the action latent variable $z$. Additionally, we integrated the third term of the RHS in Eq. (\ref{eq:gir_obj}) to further boost the backward action encoding. The third term minimizes the KL divergence between the expert policy distribution $\pi_E(a_t|s_t)$ and the action encoding distribution $q_\phi(\hat{a}_t|s_t, s_{t+1})$, where $\hat{a}_t=\textrm{Softmax}(z)$ is transformed from the latent variable $z$. In this way, we combine a backward action encoding model and a forward dynamics model into a single generative model. 

Note that the full objective in Eq. (\ref{eq:gir_obj}) is still a variational lower bound of the marginal likelihood $\log(p_\theta(s_{t+1}|s_t))$, which is reasonable to maximize as an objective of our   reward learning module. Typically, we have $p_\theta(s_{t+1}|z,s_t) \propto \exp(-\lambda\|\hat{s}_{t+1} - s_{t+1} \|_2^2)$, where $\hat{s}_{t+1}=decoder(\tilde{a}_t, s_t)$ is the reconstruction of $s_{t+1}$. By optimizing the objective, we improve the forward state transition and backward action encoding. Therefore, our reward learning module can better model the dynamics in the environment. During training, we use the latent variable $z$ as the intermediate action $\tilde{a}_t$.  

Unlike GAIL and VAIL which fit state-action pairs only, the forward state transition in our GIRL captures the dynamics of the MDP and generates next states; while its backward action encoding use states $s_t$ and $s_{t+1}$ to accurately infer expert's action information.  These two parts together can generate new state-action pairs for self-supervised exploration, which is critical for improving the generalization of intrinsic rewards of GIRL  on unseen state-action pairs in the environment, resulting in more effective exploration ability.

\textbf{\textit{Sampling-based reward for self-supervised exploration}}
After training, we infer intrinsic reward with the trained reward module following the inference procedure in Figure \ref{fig:gir}. Each time we first sample a latent variable $z$ from the learned \textit{encoder} by \[z=encoder(s_t, s_{t+1})\] and transform it into an action encoding $\hat{a}_t=\textrm{Softmax}(z)$. 
The Softmax transformation used here is continuous approximation of the discrete variable (i.e. action) for better model training  \cite{jang2016categorical}. 
We then achieve an intermediate action data $\tilde{a}_t$ by calculating a weighted sum of the true action $a_t$ and the action encoding $\hat{a}_t$, i.e. $\tilde{a}_t=\beta*a_t + (1-\beta)*\hat{a}_t$, where $\beta \in (0,1]$ is a positive weight. 
We use the learned \textit{decoder} to generate the reconstruction $\hat{s}_{t+1}$ from the state $s_t$ and the intermediate action $\tilde{a}_t$. The intrinsic reward is calculated as the reconstruction error between $\hat{s}_{t+1}$ and $s_{t+1}$:
\begin{equation}
\label{eq:giril_reward}
    r_t = \lambda \|\hat{s}_{t+1} - s_{t+1} \|_2^2
\end{equation}
\noindent where $\|\cdot\|_2$ denotes the ${\rm L}2$ norm, $\lambda$ is a positive scaling weight, $\hat{s}_{t+1}=decoder(\beta*a_t + (1-\beta)*\textrm{Softmax}(z),s_t)$.

{\bf Policy Optimization} Algorithm \ref{algo:main} summarizes GIRIL's full training procedure. The process begins by training a reward learning module for $E$ epochs (steps 3-6). In each training epoch, we sample a mini-batch demonstration data $\tilde{\mathcal{D}}$ with a mini-batch size of $B$ and maximize the objective in Eq. (\ref{eq:gir_obj}). Then in steps 7-9, we update the policy $\pi$ via any policy gradient method, e.g., PPO \cite{schulman2017proximal}, so as to optimize the policy $\pi$ with the intrinsic reward $r_t$ inferred by Eq. (\ref{eq:giril_reward}).

\begin{algorithm}[t]
\caption{Generative Intrinsic Reward driven Imitation Learning (GIRIL)}
\label{algo:main}
\begin{algorithmic}[1]
    \STATE {\bfseries Input:} Expert demonstration $\mathcal{D}=\{(s_i, a_i, s_{i+1})\}_{i=1}^N$.
    \STATE Initialize policy $\pi$, \textit{encoder} $q_\phi$ and \textit{decoder} $p_\theta$.
    \FOR{$e=1,\cdots, E$}
      \STATE Sample a batch of demonstration $\mathcal{\tilde{D}} \sim \mathcal{D}$.
      \STATE Train $q_\phi$ and $p_\theta$ to maximize the objective (\ref{eq:gir_obj}) on $\mathcal{\tilde{D}}$. 
    \ENDFOR
    \FOR{$i=1,\cdots,\textrm{MAXITER}$}
      \STATE Update policy via any policy gradient method, e.g., PPO on the intrinsic reward inferred by Eq. (\ref{eq:giril_reward}). 
    \ENDFOR
    \STATE {\bfseries Output:} Policy $\pi$.
\end{algorithmic}
\end{algorithm}

\textbf{A family of reward function for exploration}
Our method generates a family of  reward functions instead of a fixed one. After training,  we  achieve  a family of reward functions $r(s_t,a_t,\boldsymbol{z})$ with $\boldsymbol{z} \sim \mathcal{N}(\boldsymbol{\mu},\boldsymbol{\sigma})$ where $\mathcal{N}$ denotes the Gaussian distribution. The mean $\boldsymbol{\mu}$ and variance $\boldsymbol{\sigma}$ are the output of the encoding network, which is adaptively computed according to the state of environment.   The reward inference procedure is shown in Figure \ref{fig:gir}. This provides more flexibility than a fixed reward function for policy learning.  With the family of reward functions, 
the agent can perform more diverse self-supervised exploration in the environment, which is critical for the agent achieving better-than-expert performance from the limited data in a one-life demonstration.

\begin{table}[t]
    \centering
    \caption{Demonstration lengths in the Atari environment.} 
    \scalebox{1.0}{
    \begin{tabular}{c|cc|c}
    \hline
              & \multicolumn{2}{c|}{Demonstration Length} & \# Lives \\ \cline{2-3}
         Game &  One-life & Full-episode & available \\ \hline
         Space Invaders & 697 & 750 & 3 \\ 
         Beam Rider & 1,875 & 4,587 & 3 \\ 
         Breakout & 1,577 & 2,301 & 5 \\
         Q*bert & 787 & 1,881 & 4 \\
         Seaquest & 562 & 2,252 & 4 \\
         Kung Fu Master & 1,167 & 3,421 & 4 \\ \hline
    \end{tabular}
    \label{tab:demo_len}
    }
\end{table}

\begin{table*}[t]
    \centering
    \caption{Architectures of \textit{encoder}, \textit{decoder} and policy network for Atari games. } \label{table:network}
    \scalebox{1.0}{
    \begin{tabular}{c|c|c}
         \hline
        \textit{encoder} & \textit{decoder} & policy network \\ \hline
        $4\times84\times84$ States and Next States & One-hot Actions and $4\times84\times84$ States & $4\times84\times84$ States  \\ \hline
         & dense $\#$ Actions $\rightarrow$ 64, LeakyReLU &  \\
         concatenate States and Next States & dense 64 $\rightarrow$ 1024, LeakyReLU & \\
         $3 \times 3$ conv, 32 LeakyReLU & $3 \times 3$ deconv, 64 LeakyReLU & $8 \times 8$ conv, 32, stride 4, ReLU \\
         $3 \times 3$ conv, 32 LeakyReLU & $3 \times 3$ deconv, 64 LeakyReLU & $4 \times 4$ conv, 64, stride 2, ReLU \\
         $3 \times 3$ conv, 64 LeakyReLU & $3 \times 3$ deconv, 32 LeakyReLU & $3 \times 3$ conv, 32, stride 1, ReLU \\
         $3 \times 3$ conv, 64 LeakyReLU & $3 \times 3$ deconv, 32 LeakyReLU & dense $32 \times 7 \times 7$ $\rightarrow$ 512 \\
         dense 1024 $\rightarrow$ $\boldsymbol{\mu}$, dense 1024 $\rightarrow$ $\boldsymbol{\sigma}$ & concatenate with States & Categorical Distribution\\
         reparameterization $\rightarrow$ $\#$ Actions & $3 \times 3$ conv, 32 LeakyReLU & \\\hline
         Latent Variable & $4\times84\times84$ Predicted Next States & Actions \\ \hline  
    \end{tabular}
    \label{tab:atari_arch}
    }
\end{table*}

\section{Experiments and Results}
\label{sect:experiments}

\subsection{Atari}
We evaluate our proposed GIRIL on one-life demonstration data for six Atari games within OpenAI Gym \cite{brockman2016openai}. The games and demonstration details are provided in Table \ref{tab:demo_len}.

As mentioned, a one-life demonstration only contains the states and actions performed by a expert player until they die for the first time in a game. In contrast, one full-episode demonstration contains states and actions after the expert player losing all available lives in a game. Therefore, the one-life demonstration data is (much) more limited than an one full-episode demonstration. We have defined the performance tries as: basic one-life demonstration-level - gameplay up to one life lost (``one-life"), expert-level - gameplay up to all-lives lost (``one full-episode"), and beyond expert - ``better-than-expert" performance. Our ultimate goal is to train an imitation agent that can achieve better-than-expert performance from the demonstration data recorded up to losing their first life.

\subsubsection{Demonstrations}
To generate one-life demonstrations, we trained a Proximal Policy Optimization (PPO) \cite{schulman2017proximal} agent with the ground-truth reward for 10 million simulation steps. We used the PPO implementation with the default hyper-parameters in the repository \cite{kostrikov2018repo}. As Table \ref{tab:demo_len} shows, the one-life demonstrations are all much shorter than the full-episode demonstrations, which makes for extremely limited training data.

\subsubsection{Experimental Setup}

Our first step was to train a reward learning module for each game on the one-life demonstration. The proposed reward learning module consists of an \textit{encoder} and a \textit{decoder}. The \textit{encoder} consists of four convolutional layers and one fully-connected layer. Each convolutional layer is followed by a batch normalization layer (BN) \cite{ioffe2015batch}. The \textit{decoder} is nearly an inverse version of \textit{encoder} without the batch normalization layer, except that the \textit{decoder} uses the deconvolutional layer and also includes an additional fully-connected layer at the top of the \textit{decoder} and a convolutional layer at the bottom. For both the \textit{encoder} and the \textit{decoder}, we used the LeakyReLU activation \cite{maas2013rectifier} with a negative slope of 0.01. Training was conducted with the Adam optimizer \cite{kingma2014adam} at a learning rate of 3e-5 and a mini-batch size of 32 for 50,000 epochs. In each training epoch, we sampled a mini-batch of data every four states. We have summarized the detailed architectures of the \textit{encoder} and the \textit{decoder} network in Table~\ref{table:network}.

To evaluate the quality of our learned reward, we used the trained reward learning module with a $\lambda$ of 1.0 to produce rewards, and trained a policy to maximize the inferred reward function via PPO. We normalize $s_{t+1}$ and $\hat{s}_{t+1}$ to [-1,1] before calculating the reward. To further speed up the learning of value function, we performed standardization on the rewards by dividing the intrinsic rewards with a running estimate of the standard deviation of the sum of discounted rewards \cite{burda2018large}. The same discount factor $\gamma$ of 0.99 was used throughout the paper. We set $\alpha=100$ for training our reward learning module on Atari games. More experiments have been shown in the Appendix. Additionally, an ablation study in Section \ref{sect:ablation} shows the impact of standardization. 

We trained the PPO on the learned reward function for 50 million simulation steps to obtain our final policy. The PPO is trained with a learning rate of 2.5e-4, a clipping threshold of 0.1, an entropy coefficient of 0.01, a value function coefficient of 0.5, and a GAE parameter of 0.95 \cite{schulman2015high}. We compared game-play performance by our GIRIL agent against behavioral cloning (BC), and two state-of-the-art inverse reinforcement learning methods, GAIL \cite{ho2016generative} and VAIL \cite{peng2018variational}. Additionally, we adopt the intrinsic curiosity module (ICM) \cite{pathak2017curiosity,burda2018large} as a reward learning module, and also compare against with the resulting imitation learning algorithm CDIL. We have shown more details about the CDIL algorithm in the Appendix. 

\begin{figure*}[!h]
\centering\stackunder{\includegraphics[width=1.0\textwidth]{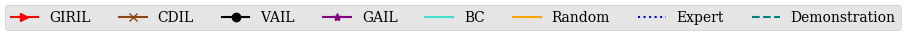}}{}
\subfigure[Space Invaders.]
{\includegraphics[width=0.33\textwidth]{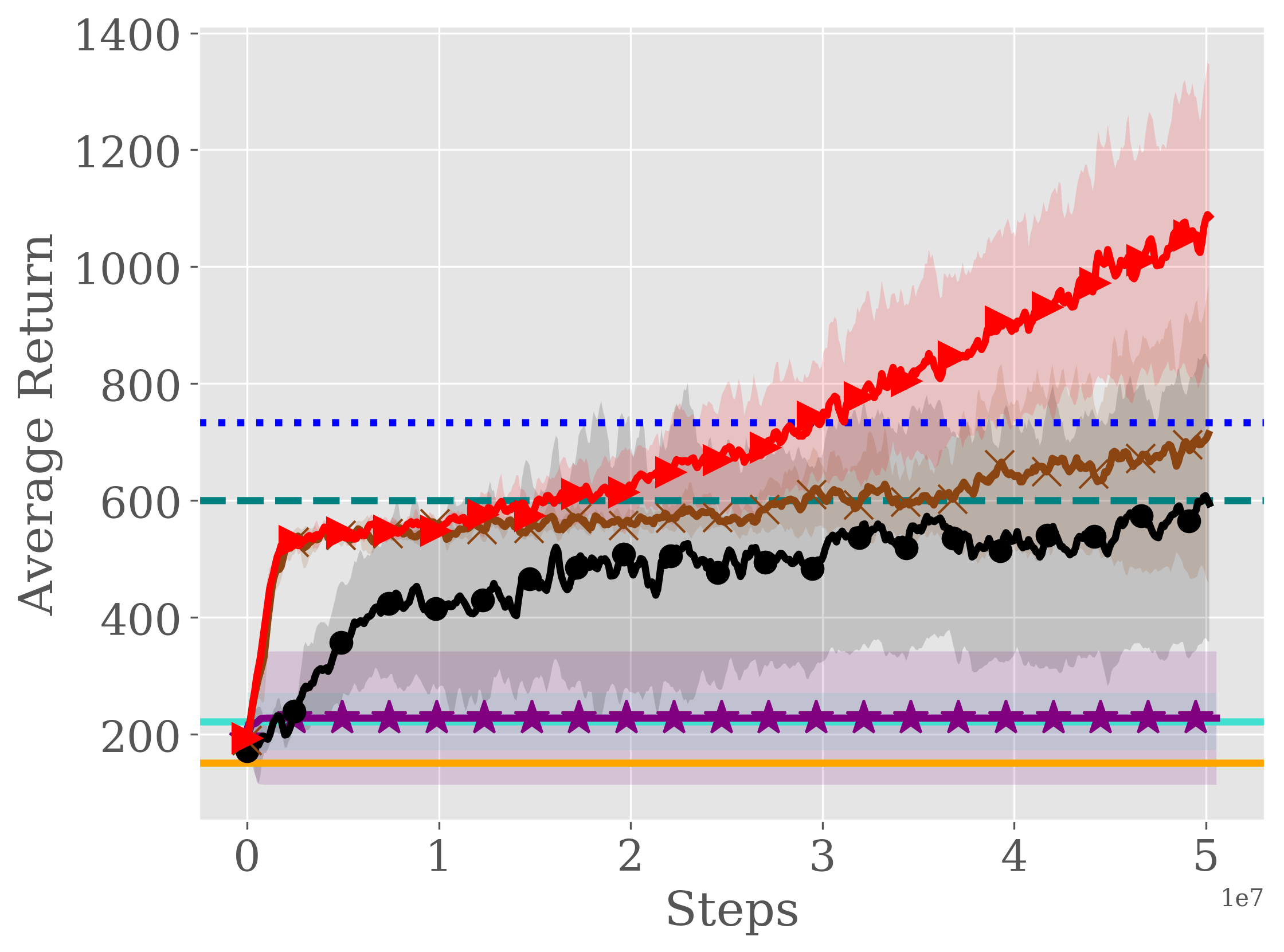}}
\subfigure[Beam Rider.]
{\includegraphics[width=0.33\textwidth]{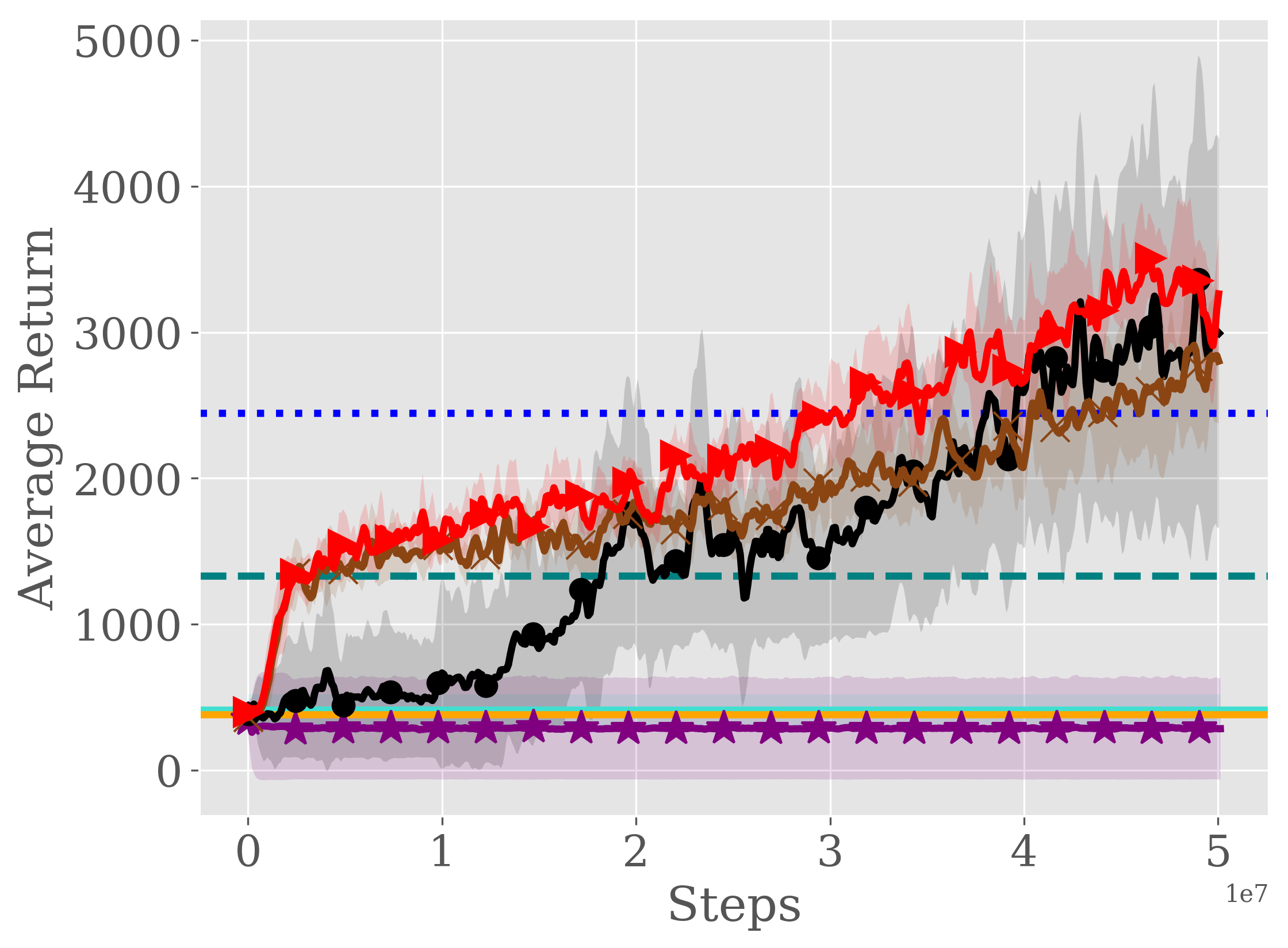}}
\subfigure[Breakout.]
{\includegraphics[width=0.33\textwidth]{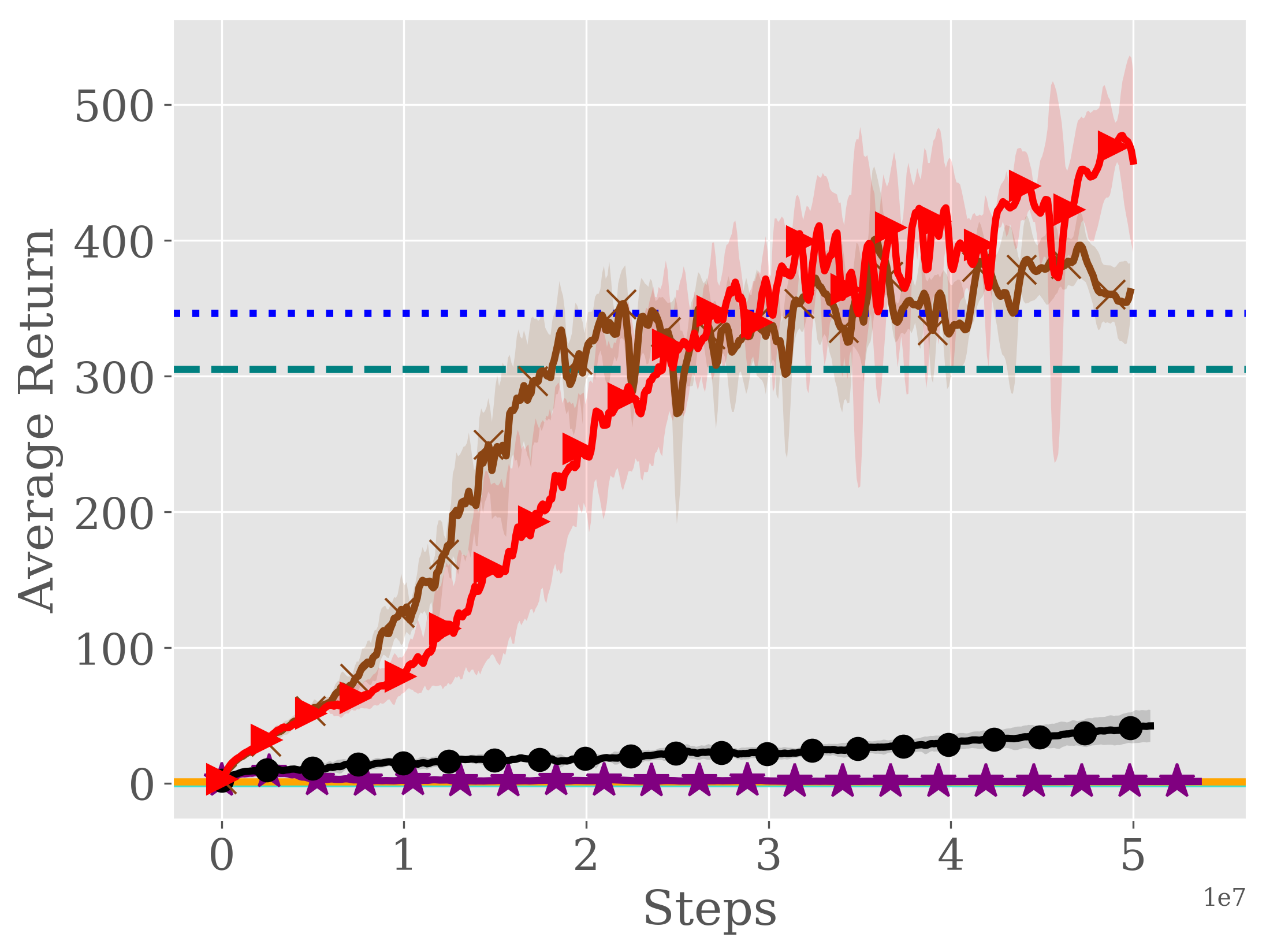}}
\subfigure[Q*bert.]
{\includegraphics[width=0.33\textwidth]{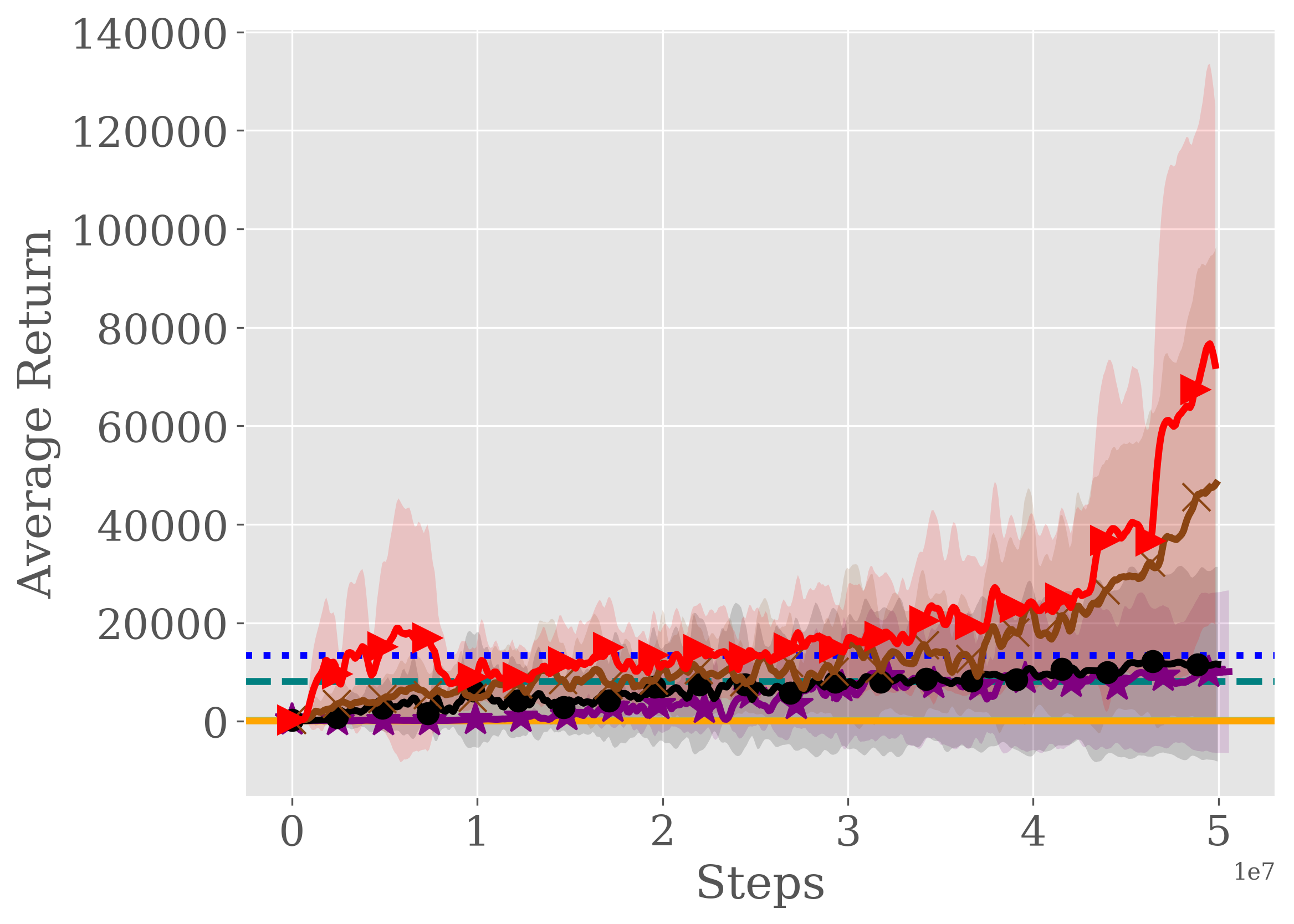}}
\subfigure[Seaquest.]
{\includegraphics[width=0.33\textwidth]{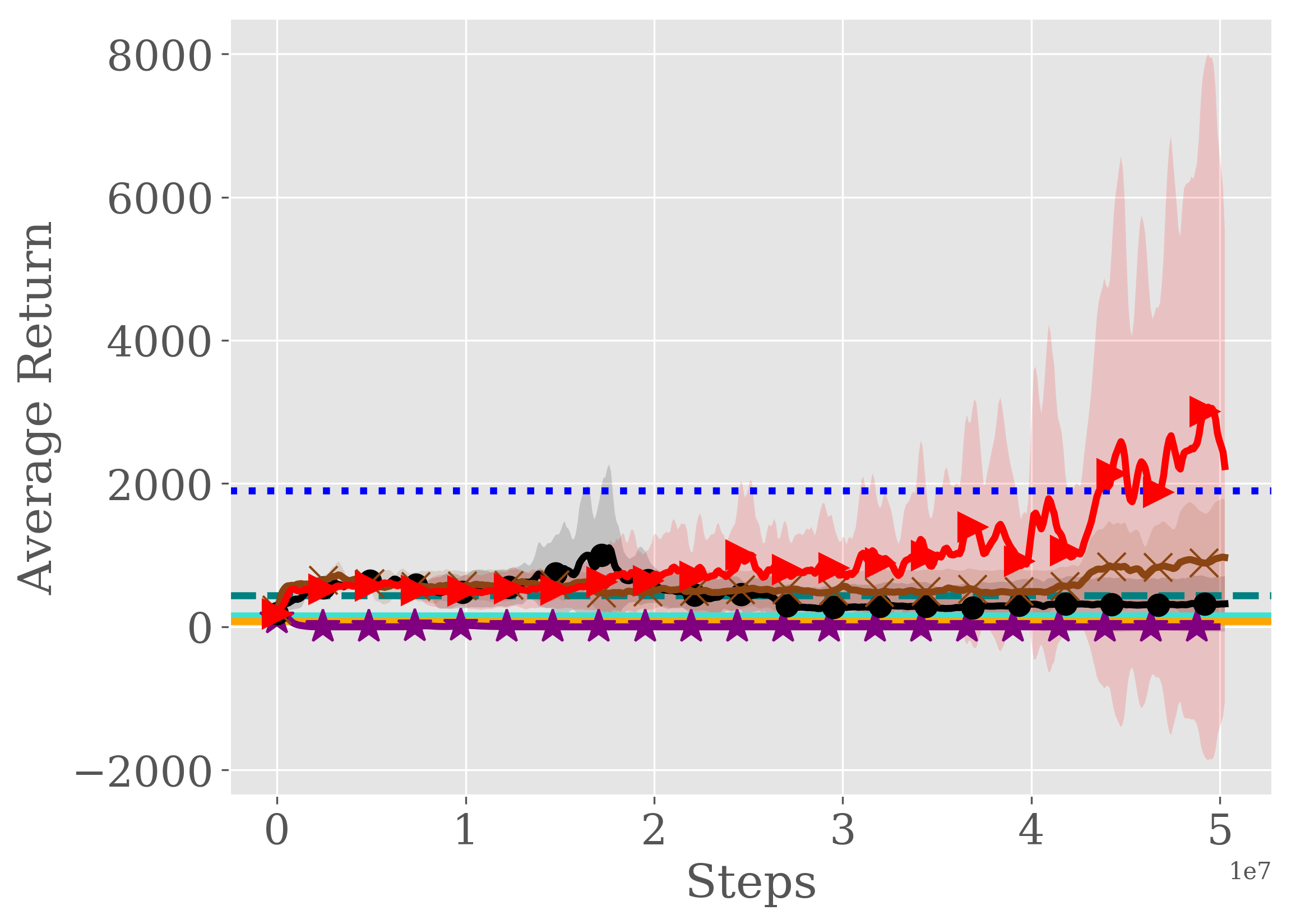}}
\subfigure[Kung Fu Master.]
{\includegraphics[width=0.33\textwidth]{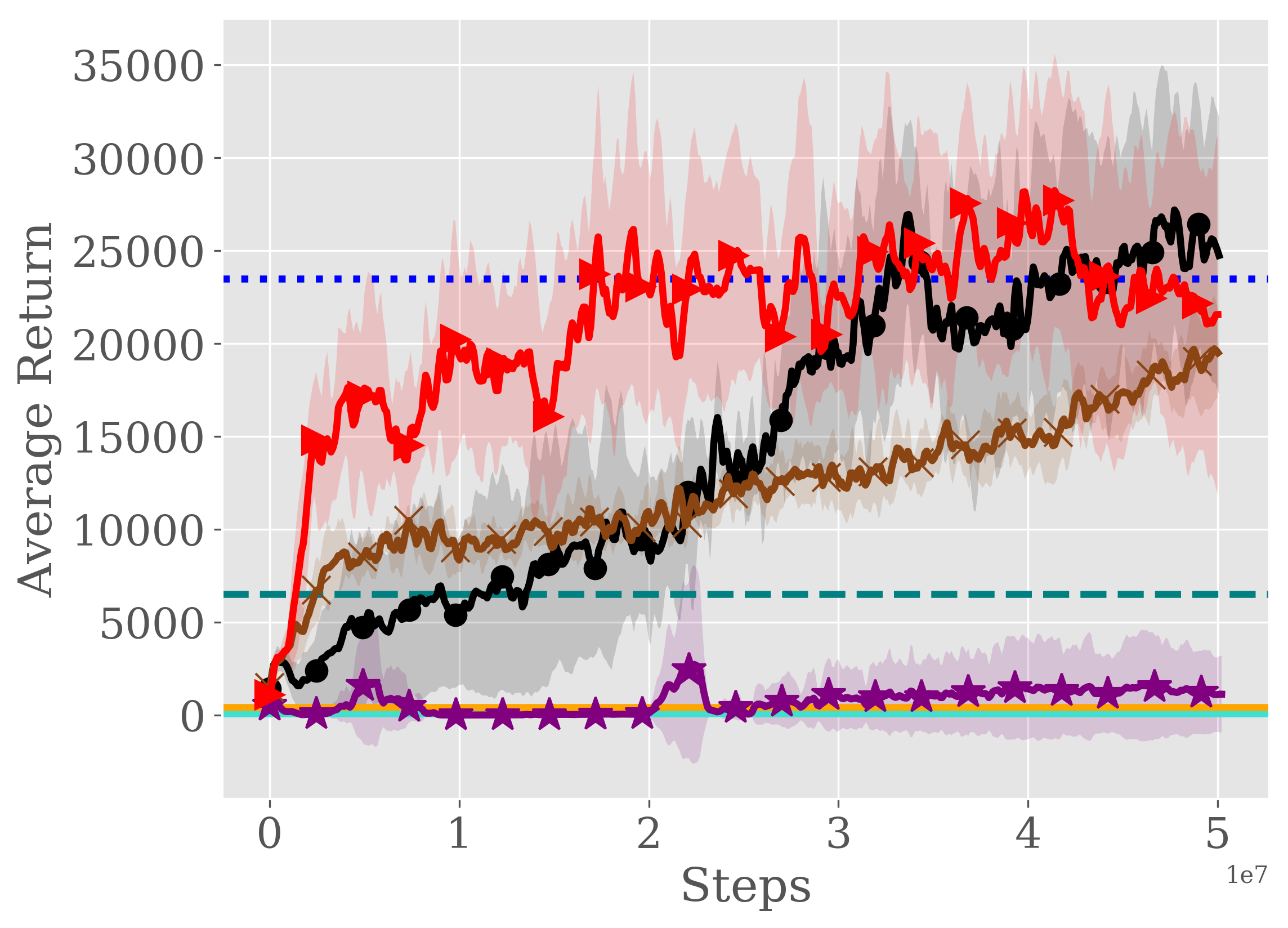}}
\caption{Average return vs. number of simulation steps on Atari games. The solid lines show the mean performance over five random seeds. The shaded area represents the standard deviation from the mean. The blue dotted line denotes the average return of expert. The area above the blue dotted line means performance beyond the expert.}
\label{fig:atari_curves}

\end{figure*}

For a fair comparison, we used an identical policy network for all methods. The architecture of the policy network is shown in the third column of Table~\ref{table:network}. We used the actor-critic approach in the PPO training for all imitation methods except BC \cite{kostrikov2018repo}. The \textit{discriminator} for both GAIL and VAIL takes in a state (a stack of four frames) and an action (represented as a 2d one-hot vector with a shape of ($|\mathcal{A}| \times 84 \times 84$), where $|\mathcal{A}|$ is the number of valid discrete actions in each environment) \cite{brown2019ranking}. The network architecture of GAIL's \textit{discriminator} $D$ is almost the same as the \textit{encoder} of our method, except that it only outputs a binary classification value, and $-\log(D(s, a))$ is the reward. VAIL was implemented following the repository of \citet{karnewar2018repo}. The \textit{discriminator} network architecture has an additional convolutional layer (with a kernel size of 4) as the final convolutional layer to encode the latent variable in VAIL. We used the default setting of 0.2 for the information constraint \cite{karnewar2018repo}. PPO with the same hyper-parameters was used to optimize the policy network for all the methods. For both GAIL and VAIL, we trained the \textit{discriminator} using the Adam optimizer with a learning rate of 0.001. The \textit{discriminator} was updated every policy step. 
The ablation study in Section \ref{sect:ablation} will show the effects of different reward functions in GAIL and different information constraints in VAIL.

\begin{table*}[t]
    \centering
    \caption{Average return of GIRIL, CDIL, BC and state-of-the-arts IRL algorithms, GAIL \cite{ho2016generative} and VAIL \cite{peng2018variational}, with one-life demonstration data on six Atari games. The results shown are the mean performance over 5 random seeds with better-than-expert performance in bold.}
    \scalebox{1.0}{
    \begin{tabular}{c|c|c|ccccc|c}
    \hline
              & Expert  & Demonstration & \multicolumn{5}{c|}{Imitation Learning Algorithms} & Random \\ 
         \cline{2-9}
         Game & Average  & Average & GIRIL & CDIL & VAIL & GAIL & BC & Average \\ \hline
         Space Invaders & 734.1 & 600.0 & \textbf{992.9} & 668.9 & 549.4 & 228.0 & 186.2 & 151.7 \\
         Beam Rider  & 2,447.7 & 1,332.0 & \textbf{3,202.3} & \textbf{2,556.9} & \textbf{2,864.1} & 285.5 & 474.7 & 379.4 \\
         Breakout & 346.4 & 305.0 & \textbf{426.9} & \textbf{369.2} & 36.1 & 1.3 & 0.9 & 1.3 \\
         Q*bert & 13,441.5 & 8,150.0 & \textbf{42,705.7} & \textbf{30,070.8} & 10,862.3 & 8,737.4 & 298.4 & 159.7 \\
         Seaquest & 1,898.8 & 440.0 & \textbf{2,022.4} & 897.7 & 312.9 & 0.0 & 155.2 & 75.5 \\
         Kung Fu Master & 23,488.5 & 6,500.0  & \textbf{23,543.6} & 17,291.6 & \textbf{24,615.9} & 1,324.5 & 44.9 & 413.7 \\ \hline
    \end{tabular}
    \label{tab:atari_performance}
    }
\end{table*}

\begin{figure}[ht]
    \centering
    \includegraphics[width=0.48\textwidth]{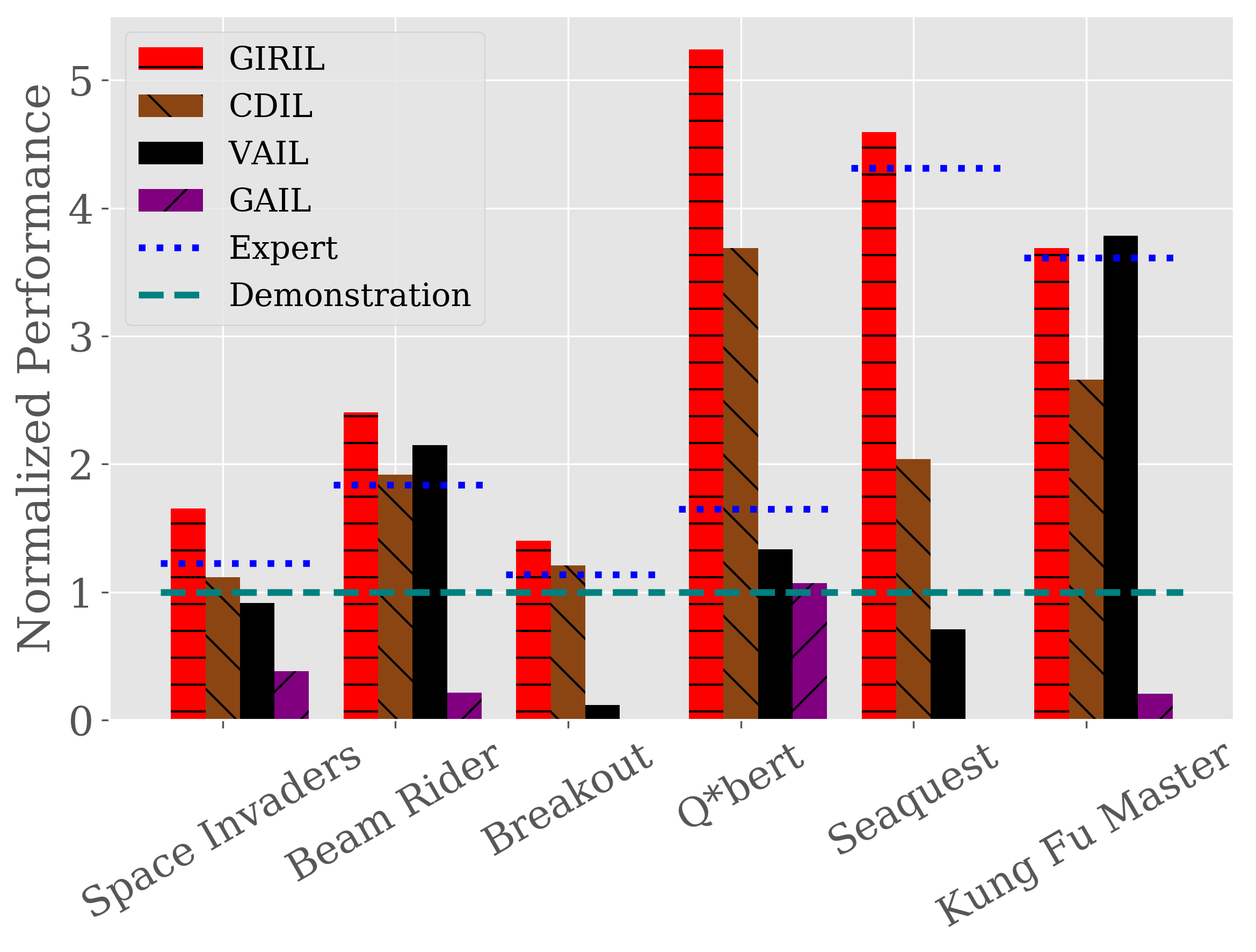}
    \caption{Performance improvement of GIRIL on six Atari games. The results are averages over 5 random seeds and reported by normalizing the one-life demonstration performance to 1.}
    \label{fig:atari_bar_plot}
\end{figure}

\subsubsection{Results}

Figure \ref{fig:atari_curves} shows the average performance of the expert, demonstration and imitation learning methods with 5 random seeds. The results reported for all games with our method, with the exception of Seaquest, were obtained by offering the standardized intrinsic reward. The Seaquest results were obtained with the original reward to show that our method also works well without standardization, achieving better performance than other baselines on multiple Atari games. There is more discussion on the influence of standardizing rewards in the ablation study in Section \ref{sect:ablation}. 

What we can clearly see from Figure \ref{fig:atari_curves} is that performance of the one-life demonstration is much lower than that of the expert. IL from such little data in a one-life demonstration is challenging. However, as shown in Figure \ref{fig:core}, our first goal is to train an agent to outperform the player's performance up to the point of losing one life. This is a start, but it is still a long way from building an agent that can ultimately succeed in the game. Therefore, the second goal is to equal the player's performance across all lives. And, since we always want a student to do better than their master, the third goal is for the agent to outperform the expert player. 

\textbf{Better-than-expert Imitation} 
Figure \ref{fig:atari_curves} shows that BC and random agent are hopelessly far from achieving the first goal. GAIL only manages to exceed the one-life demonstration for one game, Q*bert. VAIL did better than GAIL, achieving the first goal with three games and the second goal with two. CDIL managed to exceed one-life performance in all six games and even up to goal three, better-than-expert performance, on three games, while our GIRIL accomplished all three goals on all six games and often with a performance much higher than the expert player. A detailed quantitative comparison can be found in Table \ref{tab:atari_performance}. 

The full comparison of imitation performance is shown in Figure \ref{fig:atari_bar_plot}. To make the comparison more clear, we report the performance by normalizing the demonstration performance to 1. GIRIL's performance excels the expert performance on all six games, and often beating the one-life demonstration by a large margin, for example, 2 times better on Beam Rider, 3 times better on Kung Fu Master, and 4 times better on Seaquest. More impressively, our GIRIL exceeds the one-life demonstration performance on Q*bert by more than 5 times and the expert performance by more than 3 times. 

Overall, CDIL came in second our GIRIL. It outdid the one-life demonstration on all six games, but only the expert performance on three, Beam Rider, Breakout and Q*bert. It is promising to seed that both GIRIL and CDIL performed better than the two current state-of-the-arts, GAIL and VAIL. GAIL only stepped out of the lackluster performance with Q*bert. However, VAIL beat the one-life performance on Beam Rider, Q*bert and Kung Fu Master, and the expert performance on Beam Rider and Kung Fu Master. It is clear that, overall, VAIL performed better than GAIL in every game. We attribute VAIL's improvements to the use of variational information bottleneck in the discriminator \cite{tishby2015deep,alemi2016deep}.

\textbf{Comparison with generative model based imitation learning}
Although GAIL and VAIL are based on generative models just like our GIRIL, they do not have a mechanism for modeling the dynamics of the environment. Distribution matching is done by brute-force and, because one-life demonstration data can be extremely limited, direct distribution matching may result in an over-fitting problem. This is probably the most limiting factor over GAIL and VAIL's performance. Our GIRIL uses a generative model to better perform forward state transition and backward action encoding, which improves the modeling of the dynamics of MDP in environment. Moreover, our method generates a family of intrinsic reward via the sampling-based reward inference. This enables the agent to do self-supervised exploration in the environment. As shown in Figure \ref{fig:atari_curves}, GIRIL's performance improves at a sharp rate to ultimately outperform the expert player. The difference between matching demonstration performance and delivering better-than-expert performance comes as a result of the enhanced ability to explore, which is also a benefit of our generative model. A final observation is that GIRIL was often more sample-efficient than the other baselines. 

\begin{table*}[!ht]
    \centering
    \caption{Parameter Analysis of the GIRIL with different $\beta$ on Atari games. Best performance in each row is in bold.}
    \label{tab:atari_ablation_beta}
    \scalebox{1.0}{
    \begin{tabular}{c|c|c|ccccc}
    \hline
              & Expert & Demonstration &  \multicolumn{5}{c}{GIRIL with different $\beta$.} \\ 
         \cline{2-8}
         Game & Average & Average & 1.0 & 0.999 & 0.99 & 0.95 & 0.9 \\ \hline
         Space Invaders & 734.1 & 600.0 & 992.9 & \textbf{1,110.9} & 997.3 & 1,032.1 & 1,042.6 \\
         Beam Rider  & 2,447.7 & 1,332.0 & 3,202.3 & 3,351.4 & 3,276.5 & \textbf{3,402.6} &  3,145.0 \\
         Breakout & 346.4 & 305.0 & \textbf{426.9} & 397.5 & 419.3 & 416.0 & 361.5 \\
         Q*bert & 13,441.5 & 8,150.0 & \textbf{42,705.7} & 25,104.9 & 29,618.8 & 23,532.2 & 38,296.1 \\
         Seaquest & 1,898.8 & 440.0 & \textbf{2,022.4} & 526.4 & 443.3 & 433.4 & 355.6 \\
         Kung Fu Master & 23,488.5 & 6,500.0  & 23,543.6 & 16,521.4 & \textbf{23,847.7} & 20,388.5 & 19,968.6 \\ \hline
    \end{tabular}
    }
\end{table*}

\textbf{Comparison with curiosity-based reward learning}
The ICM reward learning module is also able to provide the imitation agent with an improved ability to explore. However, the limited demonstration data will give this model a tendency to overfit. Even so, CDIL exceeded the expert performance on three of the games: Beam Rider, Breakout and Q*bert. Granted, for Beam Rider and Breakout, the improvements were only negligible. However, GIRIL's use of generative model to improve dynamics modeling clearly demonstrates the performance improvements to be gained from better exploration ability.

\subsubsection{Ablation study of our method with different $\beta$. }
\label{sect:app_beta}



When $\beta=1$, we construct a basic version of our reward using the decoder only. When blending actions with different $\beta$ values, we construct a complete version of our reward that uses both encoder and decoder. Table \ref{tab:atari_ablation_beta} reports results of our methods, GIRIL, on the six Atari games with different $\beta$. The action sampling from the encoder enforces the agent with additional exploration. As a result, it potentially further improves the imitation performance versus $\beta=1$, eg., improving 1.29\% on Kung Fu Master with a $\beta$ of 0.99, improving 6.25\% on Beam Rider with a $\beta$ of 0.95 and improving 11.88\% on Space Invaders with a $\beta$ of 0.999. Overall, our method generates a family of reward functions, and enables the imitation agent to achieve better-than-expert performance on multiple Atari games. The full learning curves of our method with different $\beta$ have been shown in the Appendix.

\subsection{Continuous control tasks}
Except the above evaluation on the Atari games with high-dimensional state space and discrete action space, we also evaluated our method on continuous control tasks where the state space is low-dimensional and the action space is continuous. The continuous control tasks were from Pybullet environment \cite{coumans2019}. 

\subsubsection{Demonstrations}
To generate demonstrations, we trained a PPO agent with the ground-truth reward for 5 million simulation steps. We used PPO implementation in the repository \cite{kostrikov2018repo} with the default hyper-parameters for continuous control tasks. In each task, we used one demonstration with a fixed length of 1,000 for evaluation. 

\subsubsection{Experimental Setup}
Our first step was also to train a reward learning module for each continuous control task on the one demonstration. To build our reward learning module for continuous tasks, we used two-layer feed forward neural networks with tanh activation function as the model bases of the \textit{encoder} and \textit{decoder}. Two addition hidden layers were added to the model base in \textit{encoder} to output $\boldsymbol{\mu}$ and $\boldsymbol{\sigma}$, respectively. The dimension of latent variable $z$ is set to the action dimension for each task. Additionally, we used a two-layer feed forward neural network with tanh activation function as policy architecture. The number of hidden unit is set to 100 for all tasks. To extend our method on continuous control tasks, we made minor modification on the training objective. In Atari games, we used KL divergence to measuring the distance between the expert policy distribution and the action encoding distribution in Eq. (\ref{eq:gir_obj}). In continuous control tasks, we instead directly treated the latent variable $z$ as the action encoding and used mean squared error to measure the distance between the action encoding and the true action in the demonstration. We set the scaling weight $\alpha$ in Eq. (\ref{eq:gir_obj}) to 1.0 for all tasks. Training was conducted with the Adam optimizer \cite{kingma2014adam} at a learning rate of 3e-5 and a mini-batch size of 32 for 50,000 epochs. In each epoch, we sampled a mini-batch of data every 20 states.

To evaluate our learned reward, we used the trained reward learning module with a $\lambda$ of 1.0 to produce rewards, and trained a policy to maximize the inferred reward function via PPO. States $s_{t+1}$ and $\hat{s}_{t+1}$ were also normalized to [-1,1] before calculating rewards using Eq. (\ref{eq:cdil_reward}). We trained the PPO on the learned reward function for 10 million simulation steps to obtain our final policy. The PPO is trained with a learning rate of 3e-4, a clipping threshold of 0.1, an entropy coefficient of 0.0, a value function coefficient of 0.5, and a GAE parameter of 0.95 \cite{schulman2015high}.

\begin{table*}[!ht]
    \centering
    \caption{Average return of GIRIL, CDIL, BC and state-of-the-arts inverse reinforcement learning algorithms GAIL \cite{ho2016generative} and VAIL \cite{peng2018variational} with one demonstration data on continuous control tasks. The results shown are the mean performance over 3 random seeds with best imitation performance in bold.}
    \scalebox{0.95}{
    \begin{tabular}{c|c|ccccc}
    \hline
              & Demonstration & \multicolumn{5}{c}{Imitation Learning Algorithms}  \\ 
         \cline{2-7}
         Task & Average & GIRIL & CDIL & VAIL & GAIL & BC \\ \hline
         InvertedPendulum & 1,000.0 & \textbf{990.2} & 979.7 & 113.6 & 612.6 & 36.0 \\ 
         InvertedDoublePendulum & 9,355.1 & \textbf{9,164.9} & 7,114.7 & 725.2 & 1,409.0 & 241.6 \\ \hline
    \end{tabular}
        \label{tab:pybullet_performance}
        }
\end{table*}

For a fair comparison, we used a two-layer feed forward neural network with tanh activation function as the feature extractor in ICM, and the \textit{discriminator} in GAIL and VAIL. The number of hidden layer was also set to 100. The reward function of GAIL and VAIL was chosen according to the original papers \cite{ho2016generative,peng2018variational}. The information constraint $I_c$ in VAIL was set to 0.5 for all tasks. In our experiments with continuous control tasks, we use mean squared error as the discrepancy measure in the objective of inverse dynamics of ICM (in Eq. (\ref{eq:inverse_dynamic_obj})). To enable fast training, we trained all the imitation methods with 16 parallel processes.  

\subsubsection{Results}

\begin{figure}[!h]
\centering
\hspace{20px}\stackunder{\includegraphics[width=0.48\textwidth]{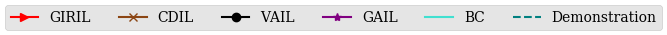}}{} \\
\subfigure[InvertedPendulum.]
{\includegraphics[width=0.235\textwidth]{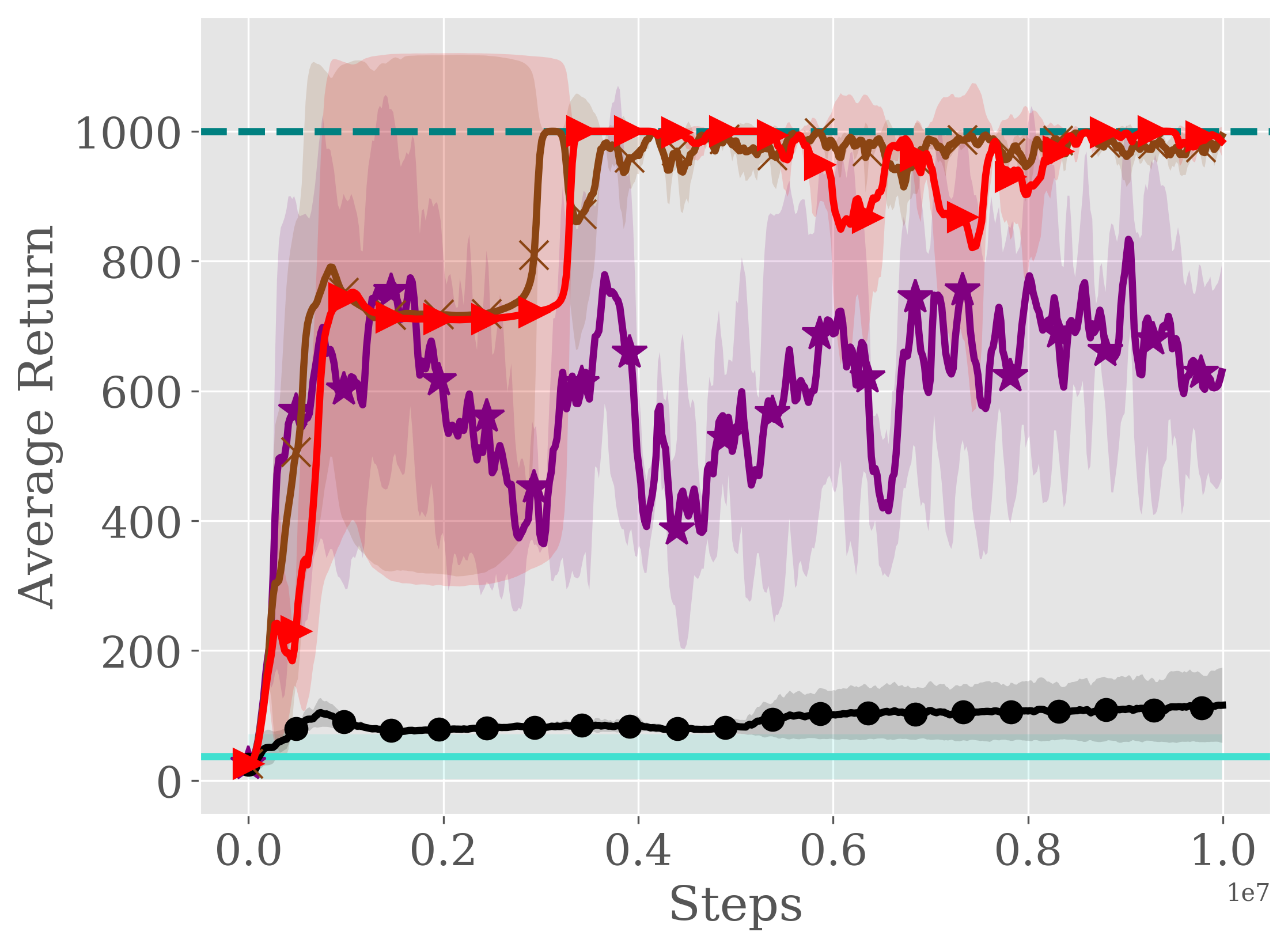}}
\subfigure[InvertedDoublePendulum.]
{\includegraphics[width=0.235\textwidth]{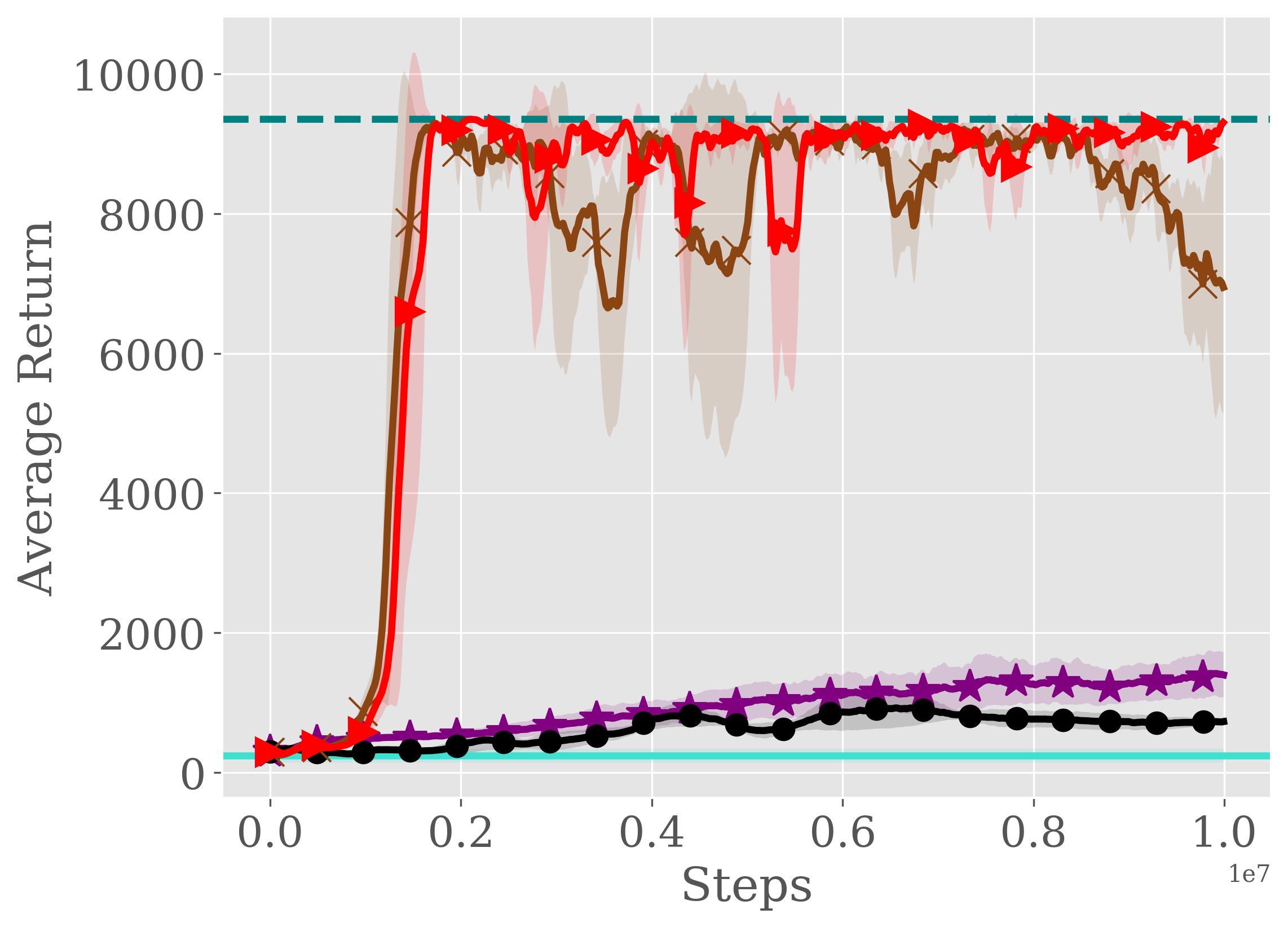}}
\caption{Average return vs. number of simulation steps on continuous control tasks. }
\label{fig:pybullet_curves}
\end{figure}

Table \ref{tab:pybullet_performance} shows the detailed quantitative comparison of the demonstration and imitation methods. The results shown in the table were the mean performance over 3 random seeds. Under a fair comparison, our method GIRIL achieves the best imitation performance in both continuous control tasks, i.e. InvertedPendulum and InvertedDoublePendulum. CDIL also achieves performance that is close to the demonstration, while still slightly worse than our method. A comparison of full learning curves can be found in Figure \ref{fig:pybullet_curves}.

\subsection{Full-episode Demonstrations}
\begin{figure}[!h]
\centering
\subfigure[Breakout.]
{\includegraphics[width=0.23\textwidth]{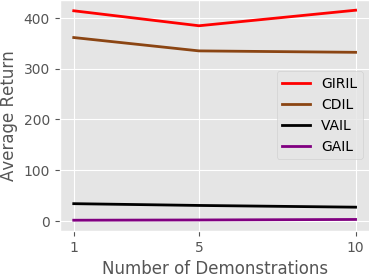}}
\subfigure[Space Invaders.]
{\includegraphics[width=0.23\textwidth]{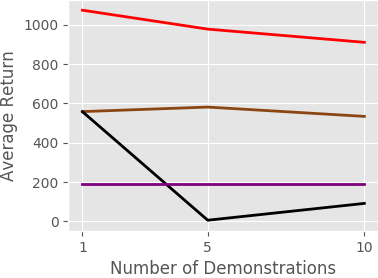}} \\
\subfigure[InvertedPendulum.]
{\includegraphics[width=0.23\textwidth]{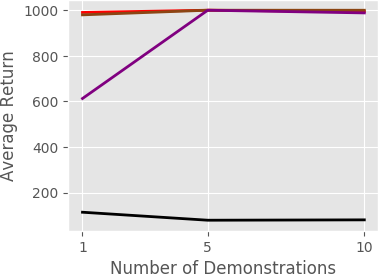}}
\subfigure[InvertedDoublePendulum.]
{\includegraphics[width=0.23\textwidth]{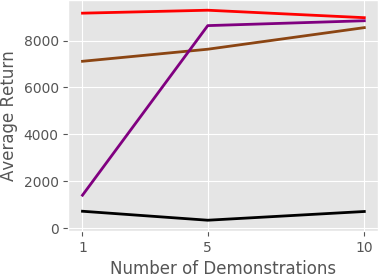}}
\caption{Average return vs. number of full-episode demonstrations on Atari games and continuous control tasks. }
\label{fig:multi_demo}
\vspace{-5px}
\end{figure}

We also evaluated our method with different number of full-episode demonstrations on both Atari games and continuous control tasks.
A comparison of average return versus number of full-episode demonstrations has been shown in Figure \ref{fig:multi_demo}. 
The results shows that our method GIRIL achieves the highest performance across different number of full-episode demonstrations on both games and tasks. CDIL usually comes the second best, and GAIL is able to achieve better performance with the increase of the demonstration number in both continuous control tasks. Detailed quantitative results have been shown in the Appendix. 

\section{Conclusion}
\label{sect:conclusion}

This paper focused on imitation learning from one-life game demonstration in the Atari environment. We propose a novel Generative Intrinsic Reward Learning (GIRL) module based on conditional VAE that combines a backward action encoding  and a forward dynamics model into one generative solution.
Our generative model can better perform forward state transition and backward action encoding, which improves the modeling of the dynamics of MDP in environment. 
The better dynamics modeling enables our model to generate more state-action pairs and more accurate rewards. Moreover, our model generates a family of intrinsic rewards, enabling the imitation agent to do sampling-based self-supervised exploration in the environment. Such exploration enables our imitation agent to learn to outperform the expert. Empirical results show that our method outperforms all other baselines including a state-of-the-art curiosity-based reward learning method, two state-of-the-art IRL methods, and behavioral cloning. 
A comparative analysis of all methods shows the advantages of our imitation learning algorithm across multiple Atari games and continuous control tasks. An interesting topic for future investigation would be to apply our GIRL to a typical, but difficult, exploration task.

\section*{Acknowledgements}
Ivor W. Tsang was supported by ARC DP180100106 and DP200101328. Xingrui Yu was supported by China Scholarship Council No. 201806450045. 


\bibliography{ref}
\bibliographystyle{icml2020}

\clearpage






\onecolumn

\appendix
\section{Appendix}
\label{sect:appendix}
\subsection{Ablation study}
\label{sect:ablation}

\subsubsection{Ablation study of our method with different $\beta$ - learning curves.}

\begin{figure*}[hbt]
\centering\stackunder{\includegraphics[width=1.0\textwidth]{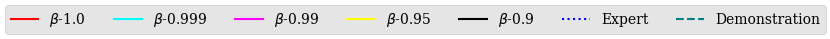}}{}
\subfigure[Space Invaders.]
{\includegraphics[width=0.33\textwidth]{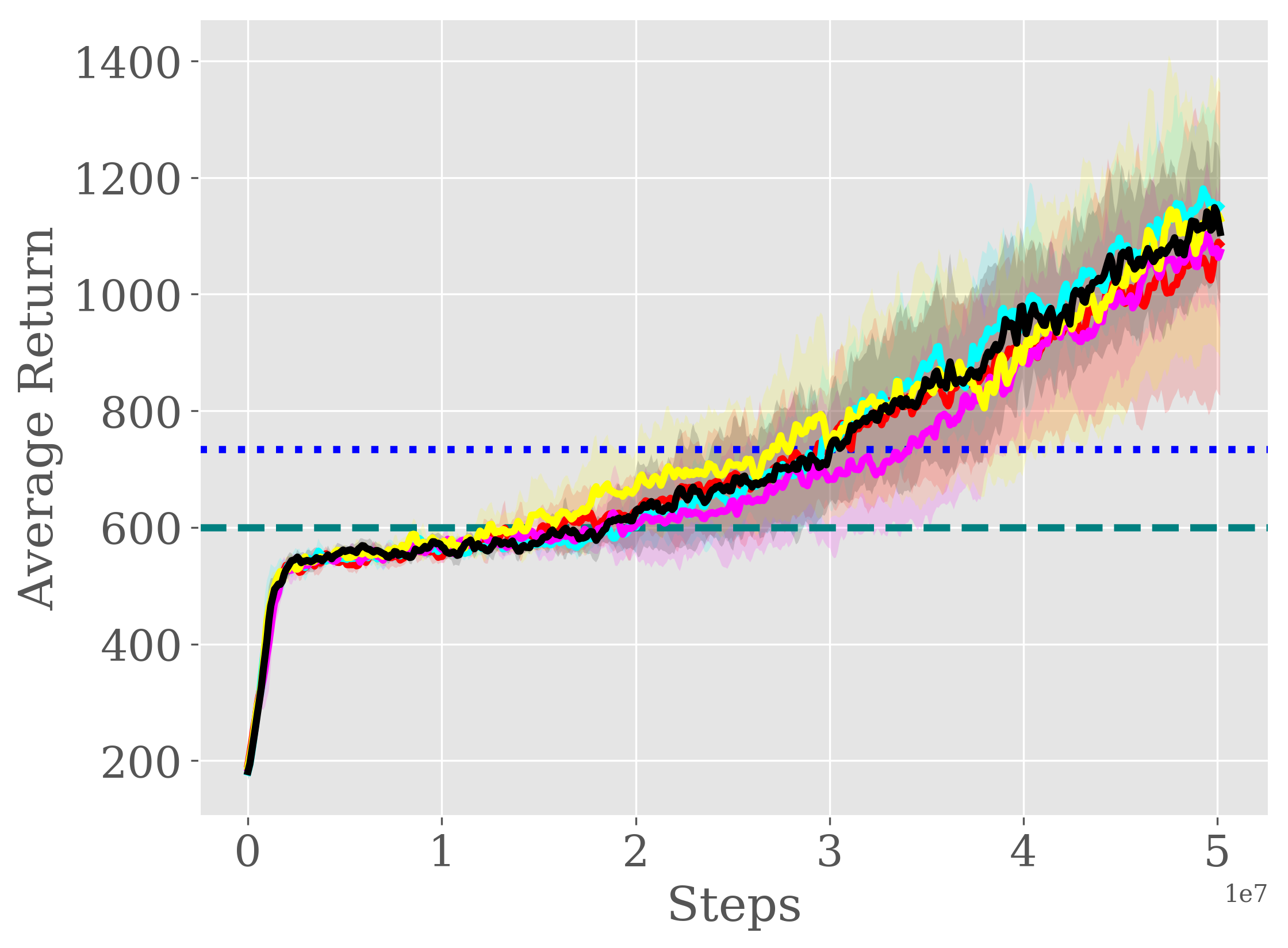}}
\subfigure[Beam Rider.]
{\includegraphics[width=0.33\textwidth]{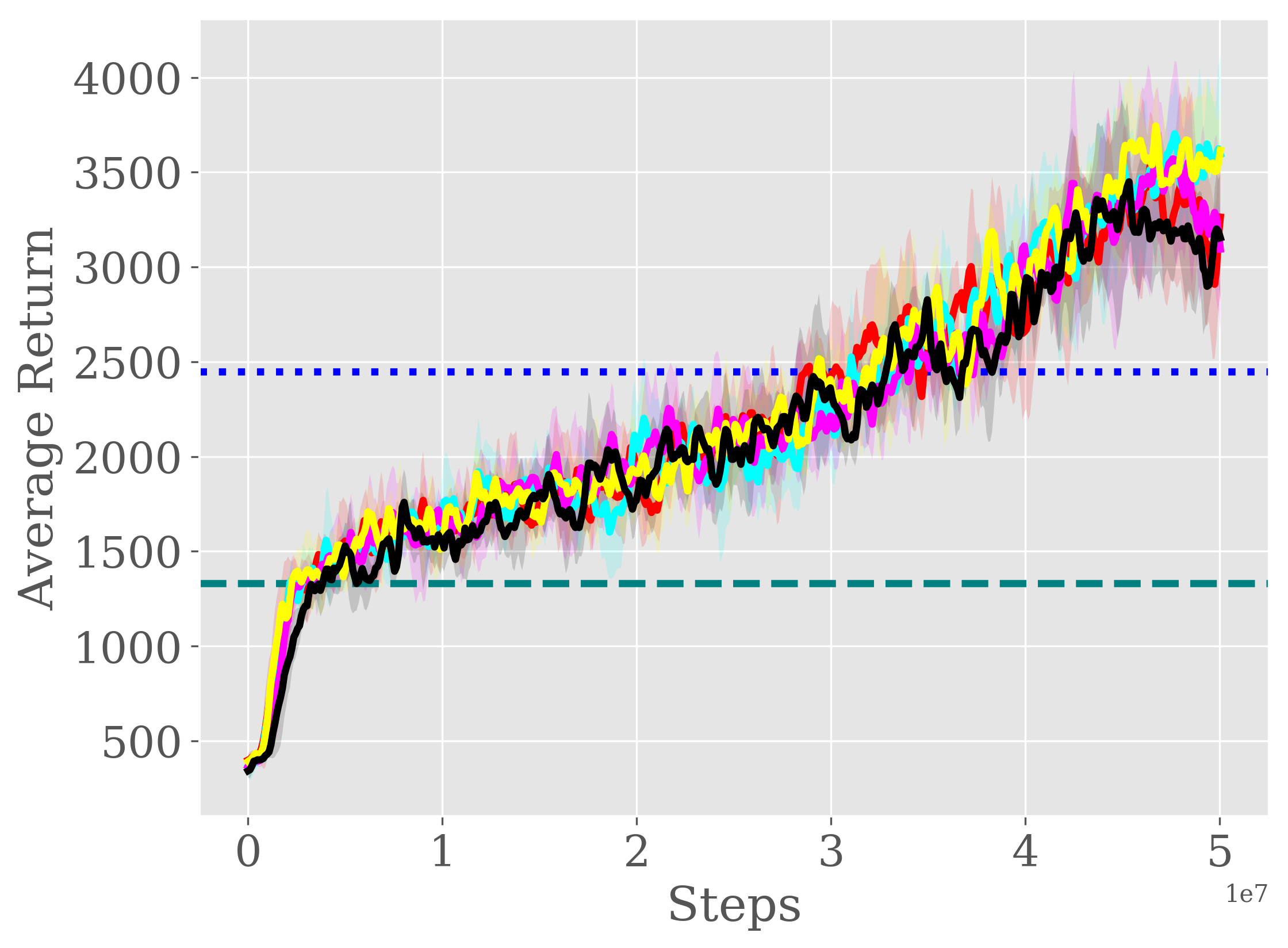}}
\subfigure[Breakout.]
{\includegraphics[width=0.33\textwidth]{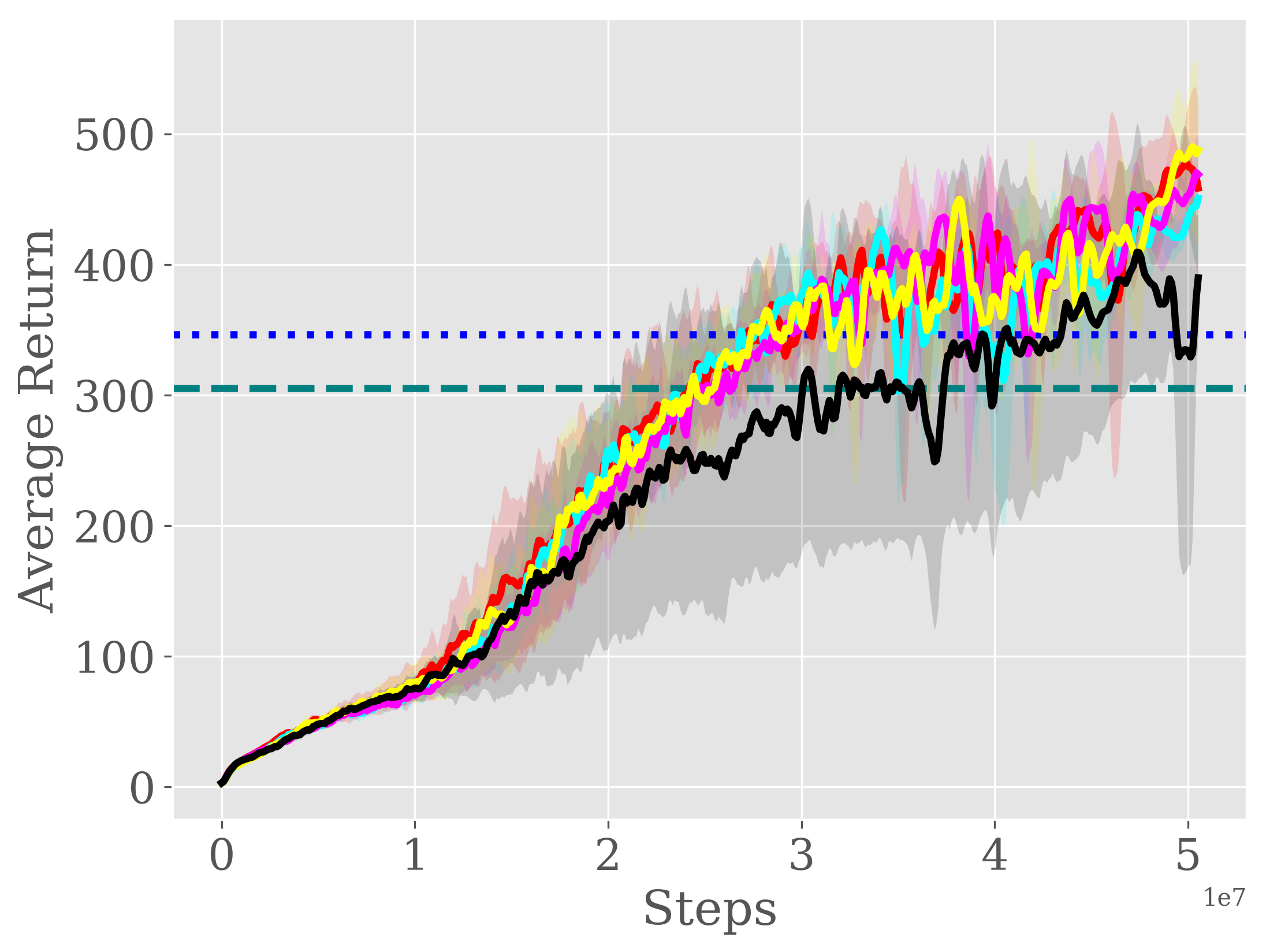}}
\subfigure[Q*bert.]
{\includegraphics[width=0.33\textwidth]{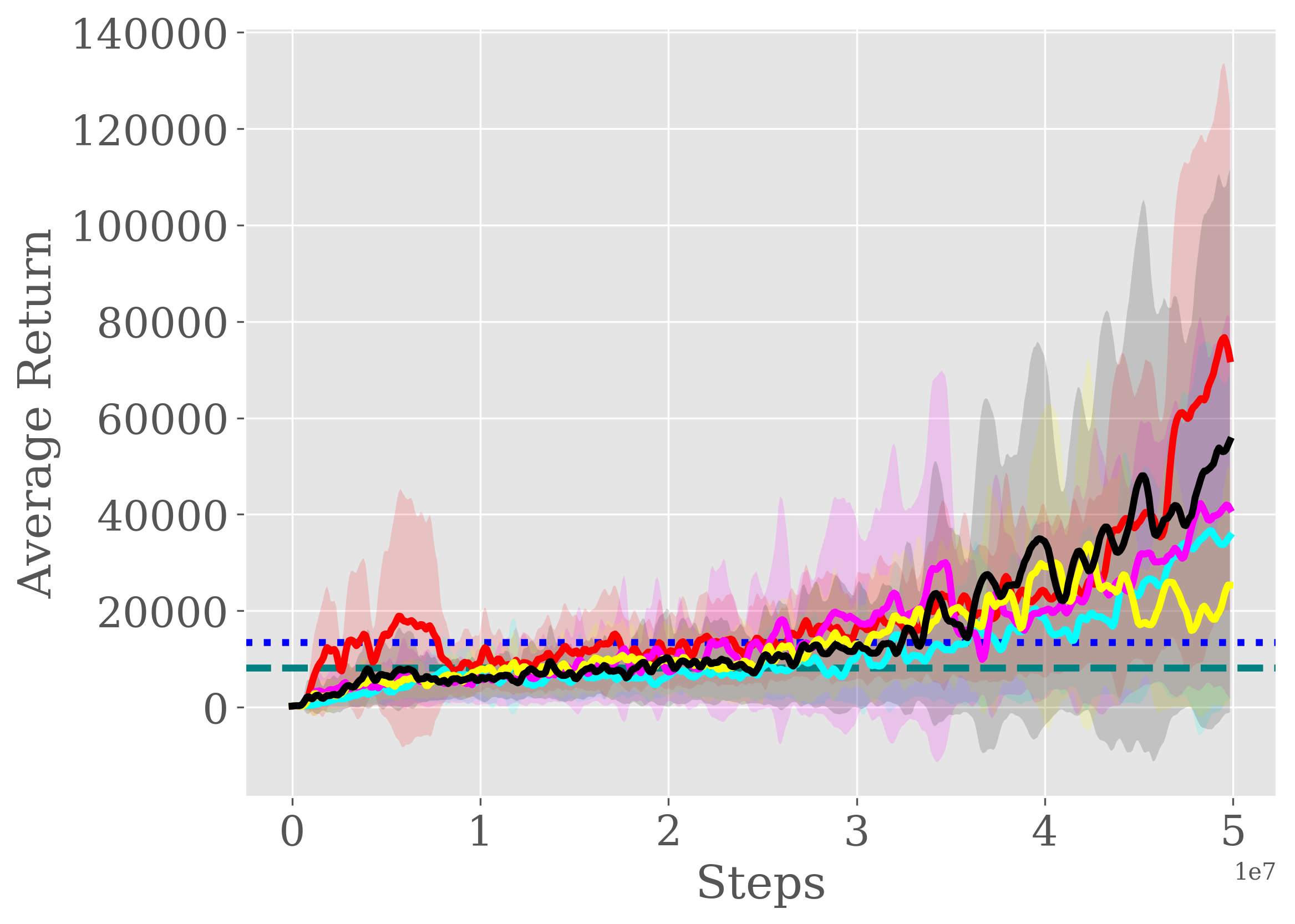}}
\subfigure[Seaquest.]
{\includegraphics[width=0.33\textwidth]{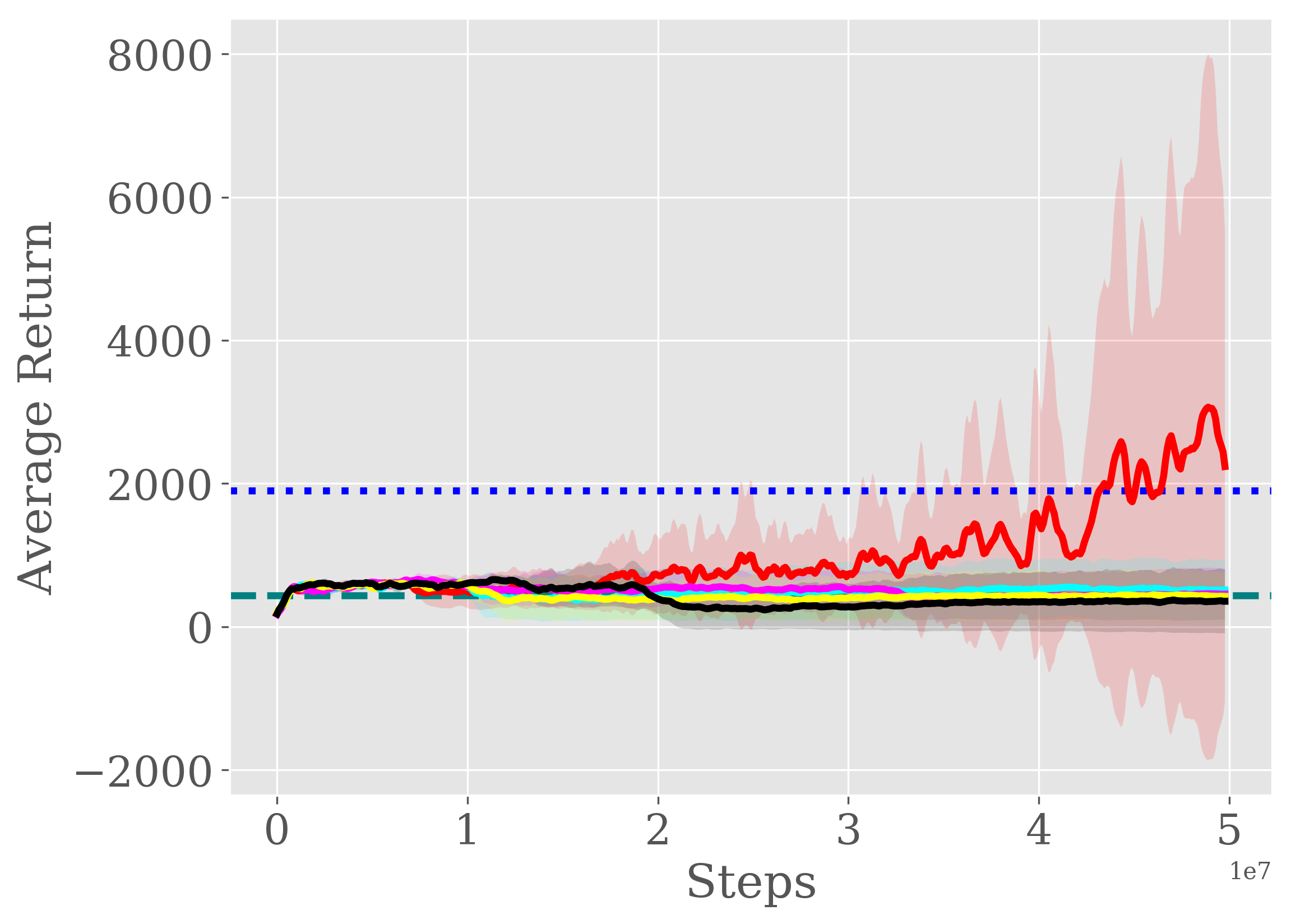}}
\subfigure[Kung Fu Master.]
{\includegraphics[width=0.33\textwidth]{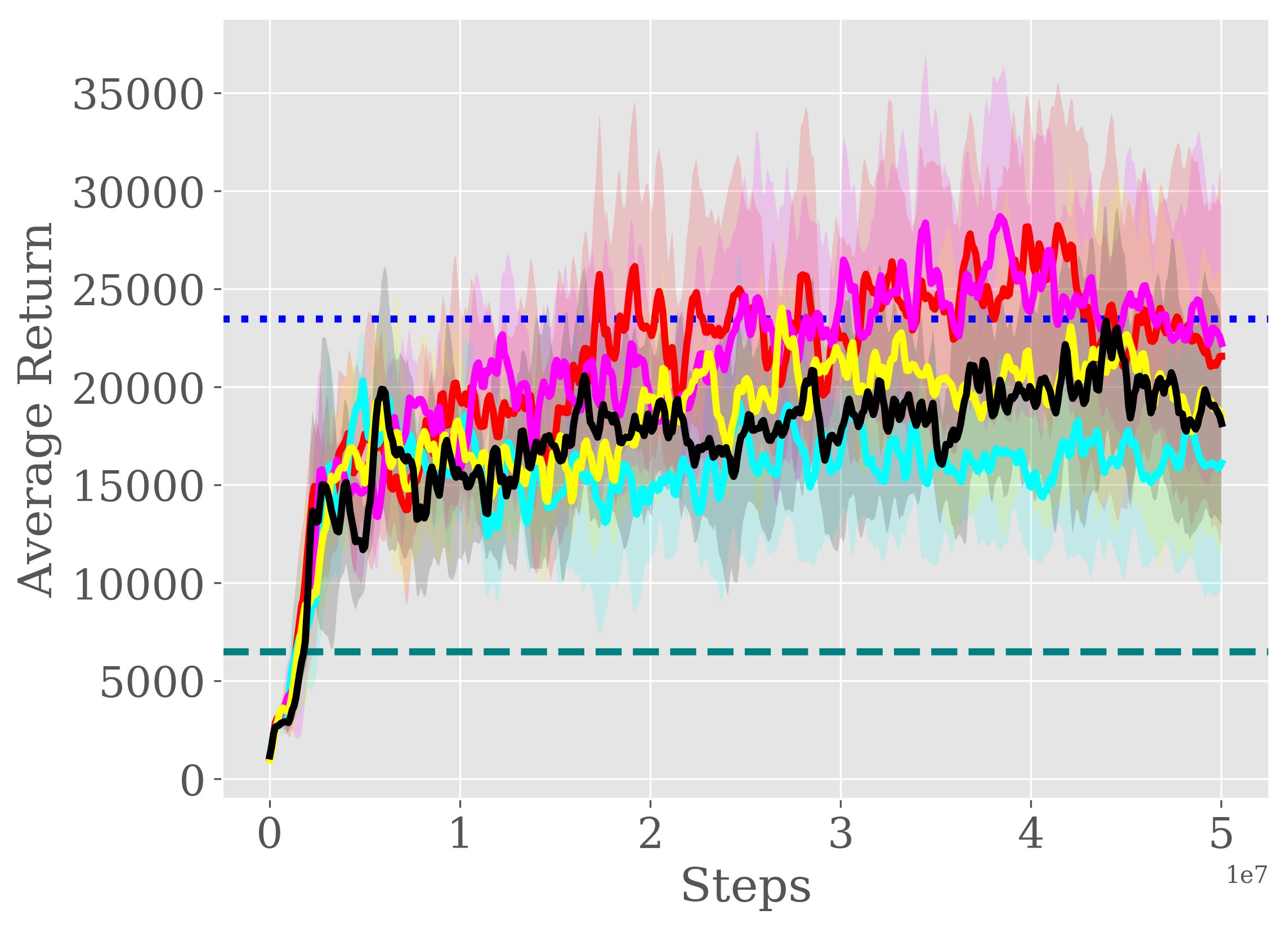}}
\caption{Average return vs. number of simulation steps on Atari games. The solid lines show the mean performance over 5 random seeds. The shaded area represents the standard deviation from the mean. The blue dotted line denotes the average return of expert. The area above the blue dotted line means performance beyond the expert.}
\label{fig:atari_ablation_beta}
\vspace{-5px}
\end{figure*}

The full learning curves of our method with different $\beta$ have been shown on Figure \ref{fig:atari_ablation_beta}.

\subsubsection{The effect of standardization in GIRIL and CDIL}

\begin{table*}[!h]
    \centering
    \caption{Ablation study of standardized intrinsic reward on the GIRIL and CDIL. The results shown are the mean performance over 5 random seeds with better-than-expert performance in bold. }
    \begin{tabular}{c|c|c|cc|cc}
    \hline
              & Expert & Demonstration & \multicolumn{2}{c|}{GIRIL} & \multicolumn{2}{c}{CDIL}  \\ 
         \cline{2-7}
         Game & Average & Average & Standardized & Original  & Standardized & Original   \\ \hline
         Space Invaders & 734.1 & 600.0 & \textbf{992.9} & 565.5 & 668.9 & 532.7  \\
         Beam Rider  & 2,447.7 & 1,332.0 & \textbf{3,202.3} & 1,810.4 & \textbf{2,556.9} & 1,808.1 \\
         Breakout & 346.4 & 305.0 & \textbf{426.9} & \textbf{375.2} & \textbf{369.2} & \textbf{369.7}  \\
         Q*bert & 13,441.5 & 8,150.0 & \textbf{42,705.7} & \textbf{21,080.3} & \textbf{30,070.8} & 12,755.4  \\
         Seaquest & 1,898.8 & 440.0 & 731.8 & \textbf{2,022.4} & 897.7 & 775.4 \\
         Kung Fu Master & 23,488.5 & 6,500.0  & \textbf{23,543.6} & \textbf{23,984.8} & 17,291.6 &  18,663.6  \\ \hline
    \end{tabular}
    \label{tab:ablation_standardization}
\end{table*}

Table \ref{tab:ablation_standardization} compares GIRIL and CDIL trained via PPO with the standardized intrinsic reward and the original intrinsic reward. With the original intrinsic reward, CDIL was able to outperform the one-life demonstration on five out of six games, but only beat the expert on Breakout. With standardization, CDIL was able to surpass the expert in two more games, Beam Rider and Q*bert. GIRIL maintain its superior performance with better-than-one-life performance on five of six games, and better-than-expert performance on four. Notably,  standardizing the reward gave GIRIL the power to outperform the one-life results with two more games and the expert results with one more game. Without standardization, GIRIL still outperformed other baselines.

\subsubsection{The effects of $r_t$ in GAIL and $I_c$ in VAIL}
We then compare GIRIL against GAIL with two different reward function $r_t$ ($r_t^{(1)}$=$-\log(D(s_t, a_t))$ and $r_t^{(2)}$=$-\log(1-D(s_t, a_t))$, where $D$ is the \textit{discriminator}) and VAIL with two different information constraints $I_c$ ($I_c$=$0.2$, and $I_c$=$0.5$). $I_c$=$0.2$ and $I_c$=$0.5$ are the default hyper-parameters in \citet{karnewar2018repo} and \citet{peng2018variational}, respectively. The results are provided in Table \ref{tab:ablation_vail_gail}. 

\begin{table*}[!h]
    \centering
    \caption{Parameter Analysis of the GIRIL versus VAIL with different information constraints $I_c$, and versus GAIL with different rewards $r_t$, i.e. $r_t^{(1)}$=$-\log(D(s_t, a_t))$ and $r_t^{(2)}$=$-\log(1-D(s_t, a_t))$. The results shown are the mean performance over 5 random seeds with better-than-expert performance in bold.}
    \begin{tabular}{c|c|c|c|cc|cc}
    \hline
              & Expert & Demonstration & GIRIL & \multicolumn{2}{c|}{VAIL ($I_c$)} & \multicolumn{2}{c}{GAIL ($r_t$)} \\ 
         \cline{2-8}
         Game & Average & Average & Average & 0.2 & 0.5 & $r_t^{(1)}$ & $r_t^{(2)}$\\ \hline
         Space Invaders & 734.1 & 600.0 & \textbf{992.9} & 549.4 &  426.5 & 228.0 & 129.9 \\
         Beam Rider  & 2,447.7 & 1,332.0 & \textbf{3,202.3} & \textbf{2,864.1}  & \textbf{2,502.7} & 285.5 & 131.3 \\
         Breakout & 346.4 & 305.0 & \textbf{426.9} & 36.1 & 27.2 & 1.3 & 2.5 \\
         Q*bert & 13,441.5 & 8,150.0 & \textbf{42,705.7} & 10,862.3 & \textbf{54,247.3} & 8,737.4 & 205.3 \\
         Seaquest & 1,898.8 & 440.0 & \textbf{2,022.4} & 312.9 & 1,746.7 & 0.0& 28.9 \\
         Kung Fu Master & 23,488.5 & 6,500.0  & \textbf{23,543.6} & \textbf{24,615.9} & 14,709.3 & 1,324.5 & 549.7 \\ \hline
    \end{tabular}
    \label{tab:ablation_vail_gail}
\end{table*}

As the results show, GAIL with $-\log(D(s_t, a_t))$ performed better than that with $-\log(1-D(s_t, a_t))$. VAIL showed similar performance no matter the information constraint. Both outplayed the expert on two games - an overall worse performance than CDIL with standardized reward and GIRIL with both types of reward.

\subsubsection{The effect of the number of full-episode demonstrations.} We also evaluated our method with different number of full-episode demonstrations on both Atari games and continuous control tasks. Table \ref{tab:ablation_num_demo_atari_breakout} and Table \ref{tab:ablation_num_demo_atari_spaceinvaders} show the detailed quantitative comparison of imitation learning methods across different number of full-episode demonstrations in the games, Breakout and Space Invaders. The comparisons on two continuous control tasks, InvertedPendulum and InvertedDoublePendulum, have been shown in Table \ref{tab:ablation_num_demo_pybullet_invertedpendulum} and Table \ref{tab:ablation_num_demo_pybullet_inverteddoublependulum}. 

The results shows that our method GIRIL achieves the highest performance across different numbers of full-episode demonstrations, and CDIL usually comes the second best. GAIL is able to achieve better performance with the increase of the demonstration number in both continuous control tasks.

\begin{table*}[!h]
    \centering
    \caption{Parameter Analysis of the GIRIL versus other baselines with different number of full-episode demonstrations on Breakout game. The results shown are the mean performance over 5 random seeds with best performance in bold.}
    \begin{tabular}{c|c|c|c|c}
    \hline
          \# Demonstrations & GIRIL & CDIL & VAIL & GAIL \\ \hline
         1 & \textbf{413.9} & 361.2 & 34.0 &  1.4 \\ \hline
         5 & \textbf{384.4} & 334.9 & 30.5 &  1.9 \\ \hline
         10 & \textbf{415.0} & 332.1 & 27.1 &  2.9 \\ \hline
    \end{tabular}
    \label{tab:ablation_num_demo_atari_breakout}
\end{table*}

\begin{table*}[!h]
    \centering
    \caption{Parameter Analysis of the GIRIL versus other baselines with different number of full-episode demonstrations on Space Invaders game. The results shown are the mean performance over 5 random seeds with best performance in bold.}
    \begin{tabular}{c|c|c|c|c}
    \hline
          \# Demonstrations & GIRIL & CDIL & VAIL & GAIL \\ \hline
         1 & \textbf{1,073.8} & 557.5 & 557.0 &  190.0 \\ \hline
         5 & \textbf{977.6} & 580.6 & 4.4 &  190.0 \\ \hline
         10 & \textbf{910.3} & 533.2 & 90.0 &  190.0 \\ \hline
    \end{tabular}
    \label{tab:ablation_num_demo_atari_spaceinvaders}
\end{table*}

\begin{table*}[!h]
    \centering
    \caption{Parameter Analysis of the GIRIL versus other baselines with different number of full-episode demonstrations on InvertedPendulum task. The results shown are the mean performance over 5 random seeds with best performance in bold.}
    \begin{tabular}{c|c|c|c|c}
    \hline
          \# Demonstrations & GIRIL & CDIL & VAIL & GAIL \\ \hline
         1 & \textbf{990.2} & 979.7 & 113.6 &  612.6 \\ \hline
         5 & \textbf{1,000.0} & 1,000.0 & 78.5 &  1,000.0 \\ \hline
         10 & 994.4 & \textbf{999.9} & 80.1 &  988.2 \\ \hline
    \end{tabular}
    \label{tab:ablation_num_demo_pybullet_invertedpendulum}
\end{table*}

\begin{table*}[!h]
    \centering
    \caption{Parameter Analysis of the GIRIL versus other baselines with different number of full-episode demonstrations on InvertedDoublePendulum task. The results shown are the mean performance over 5 random seeds with best performance in bold.}
    \begin{tabular}{c|c|c|c|c}
    \hline
          \# Demonstrations & GIRIL & CDIL & VAIL & GAIL \\ \hline
         1 & \textbf{9,164.9} & 7,114.7 & 725.2 &  1,409.0 \\ \hline
         5 & \textbf{9,290.4} & 7,628.7 & 342.9 &  8,634.5 \\ \hline
         10 & \textbf{8,972.8} & 8,548.6 & 714.8 &  8,842.0 \\ \hline
    \end{tabular}
    \label{tab:ablation_num_demo_pybullet_inverteddoublependulum}
\end{table*}


\subsection{Details of the \textit{curiosity-driven imitation learning} (CDIL)}
\label{sect:app_cdil}
\begin{figure}[hbt]
    \centering
    \includegraphics[scale=0.4]{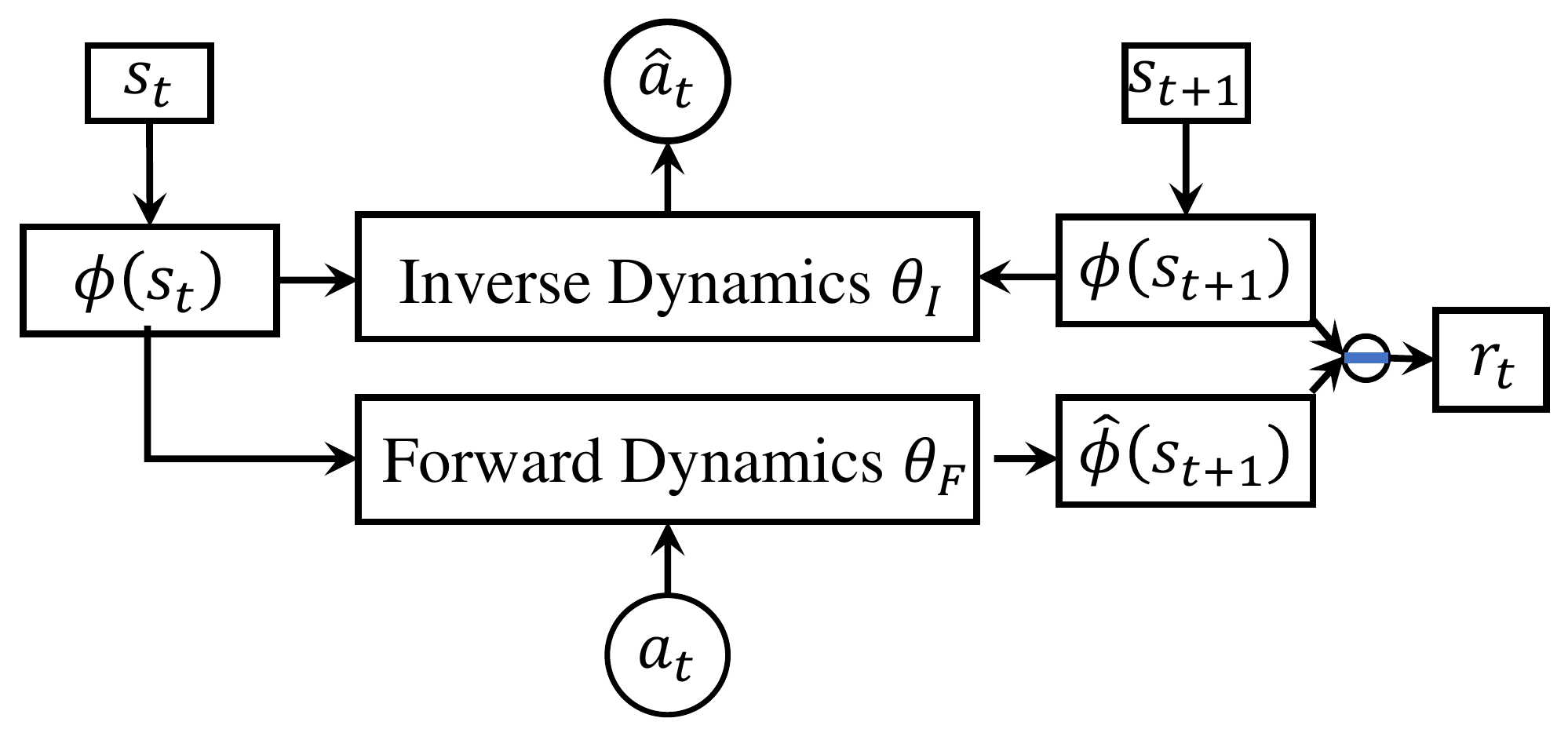}
    \caption{Intrinsic Curiosity Module (ICM).}
    \label{fig:icm}
\end{figure}

The Intrinsic Curiosity Module (ICM) is a natual choice for reward learning in imitaiton learning. ICM is a state-of-the-art exploration method \cite{pathak2017curiosity,burda2018large} that transforms high-dimensional states into a visual feature space and then impose a cross-entropy loss and a Euclidean loss to learn the features with a self-supervised inverse dynamics model. Further, the prediction error in the feature space becomes the intrinsic reward function for exploration. As illustrated in Figure \ref{fig:icm}, ICM encodes the states $s_t$, $s_{t+1}$ into features and then the inverse dynamics model $g_{\theta_I}$ is trained to predict actions from the states features $\phi(s_t)$ and $\phi(s_{t+1})$. Additionally, the forward model $f_{\theta_F}$ takes a feature $\phi(s_t)$ and an action $a_t$ as input and predicts the feature representation of state $s_{t+1}$. The intrinsic reward is calculated as the curiosity, i.e. the prediction error in the feature space.

In ICM, the inverse dynamics model is used to predict the action $\hat{a}_t=g_{\theta_I}(\phi(s_t), \phi(s_{t+1}))$, and is optimized by:
\begin{equation}
\label{eq:inverse_dynamic_obj}
    \min_{\theta_I} L_I(\hat{a}_t, a_t),
\end{equation}
\noindent where $L_I$ is the loss function measures the discrepancy between the predicted and actual action. In our experiments, we use cross-entropy loss for Atari games and mean squared error (MSE) for continuous control tasks.

The forward dynamics model estimates the feature of next state $\hat{\phi}(s_{t+1})=f_{\theta_F}(\phi(s_t), a_t)$, and is optimized by:
\begin{equation}
    \min_{\theta_F} L_F(\phi(s_t), \hat{\phi}(s_{t+1}))= \| \hat{\phi}(s_{t+1}) - \phi(s_{t+1}) \|_2^2,
\end{equation}
\noindent where $\|\cdot\|_2$ is the ${\rm L}2$ norm.

ICM is optimized by minimizing the overall objective as follows:
\begin{equation}
    \label{eq:icm_obj}
    \min_{\theta_I,\theta_F} L_I + L_F 
\end{equation}

The intrinsic reward signal $r_t$ is calculated as the prediction error in feature space:
\begin{equation}
\label{eq:cdil_reward}
    r_t = \lambda\|\hat{\phi}(s_{t+1})-\phi(s_{t+1}) \|_2^2
\end{equation}
\noindent where $\|\cdot\|_2$ is the ${\rm L}2$ norm, and $\lambda$ is a scaling weight. In all experiments, $\lambda=1$.

Thus, our solution combines ICM for reward learning and reinforcement learning. The full CDIL training procedure is summarized in Algorithm \ref{algo:cdil}.

\begin{algorithm}[hbt]
\caption{Curiosity-driven imitation learning (CDIL)}
\label{algo:cdil}
\begin{algorithmic}[1]
    \STATE {\bfseries Input:} Expert demonstration data $\mathcal{D}=\{(s_i, a_i)\}_{i=1}^N$.
    \STATE Initialize policy $\pi$, \textit{encoder} $q_\phi$ and \textit{decoder} $p_\theta$.
    \FOR{$e=1,\cdots, E$}
      \STATE Sample a batch of demonstration $\mathcal{\tilde{D}} \sim \mathcal{D}$.
      \STATE Train $f_{\theta_F}$ and $g_{\theta_I}$ to optimize the objective (\ref{eq:icm_obj}) on $\mathcal{\tilde{D}}$. 
    \ENDFOR
    \FOR{$i=1,\cdots,\textrm{MAXITER}$}
      \STATE Update policy parameters via any policy gradient method, e.g., PPO on the intrinsic reward inferred by Eq. (\ref{eq:cdil_reward}). 
    \ENDFOR
    \STATE {\bfseries Output:} Policy $\pi$.
\end{algorithmic}
\end{algorithm}

In brief, the process begins by training ICM for $E$ epochs (Steps 3-6). In each training epoch, we sample a mini-batch of demonstration data $\tilde{D}$ with a size of $B$ and maximize the objective in Eq. (\ref{eq:icm_obj}). Steps 7-9 perform policy gradient steps, e.g., PPO\cite{schulman2017proximal}, so as to optimize the policy $\pi$ with the intrinsic reward $r_t$ inferred with ICM using Eq. (\ref{eq:cdil_reward}). We treated CDIL as a related baseline in our experiments, using the feature extractor with the same architecture as the \textit{encoder} except for the final dense layer. We trained the ICM using the Adam optimizer \cite{kingma2014adam} with a learning rate of 3e-5 and a mini-batch size of 32 for 50, 000 epochs. In each training epoch, we sample a mini-batch data every four states for Atari games and every 20 states for continuous control tasks.

\clearpage

\end{document}